\documentclass[letterpaper]{article} 
\usepackage{aaai24}  
\usepackage{times}  
\usepackage{helvet}  
\usepackage{courier}  
\usepackage[hyphens]{url}  
\usepackage{graphicx} 
\usepackage{natbib}  
\usepackage{caption} 
\usepackage{algorithm}
\usepackage{tabularray}
\usepackage{graphicx}
\usepackage{subcaption}
\usepackage{pifont}
\usepackage{bbding}
\usepackage{enumitem}
\usepackage{amssymb}
\usepackage{amsthm}
\usepackage{booktabs}
\usepackage{multirow}
\usepackage[table,xcdraw]{xcolor}
\usepackage{colortbl}
\usepackage{amsopn}
\newtheorem{lemma}{Lemma}
\usepackage{amsmath}

\makeatletter
\newcommand{\stitle}[1]{\noindent\textbf{#1\@addpunct{.}}}
\makeatother

\usepackage{bm}
\def\vtheta{{\bm{\theta}}}
\def\vx{{\bm{x}}}
\def\vz{{\bm{z}}}

\def\e{t}
\def\vze{\vz_t}
\def\vxe{\vx_t}

\def\vg{{\bm{g}}}

\def\d{\mathrm{d}}

\def\X{{\bm{\mathcal{X}}}}
\def\Y{{\mathcal{Y}}}

\def\S{{\mathcal{S}}}

\def\I{I}
\def\P{P}
\def\D{D}

\DeclareMathOperator{\simi}{sim}

\def\E{{\mathcal{E}}}

\def\mH{{\bm{H}}}
\usepackage{algorithm}
\usepackage{algorithmic}

\usepackage{newfloat}
\usepackage{listings}
\DeclareCaptionStyle{ruled}{labelfont=normalfont,labelsep=colon,strut=off} 
\floatstyle{ruled}
\newfloat{listing}{tb}{lst}{}
\floatname{listing}{Listing}
\title{AIDE: Antithetical, Intent-based, and Diverse Example-Based Explanations}
\author {
    Ikhtiyor Nematov\textsuperscript{\rm 1,\rm 2},
    Dimitris Sacharidis\textsuperscript{\rm 1},
    Tomer Sagi\textsuperscript{\rm 2},
    Katja Hose\textsuperscript{\rm 3}
}
\affiliations {
    \textsuperscript{\rm 1}Universite Libre de Bruxelles, Belgium\\
    \textsuperscript{\rm 2}Aalborg University, Denmark\\
    \textsuperscript{\rm 3}TU Wien, Austria\\
    ikhtiyor.nematov@ulb.be, dimitris.sacharidis@ulb.be, tsagi@cs.aau.dk, katja.hose@tuwien.ac.at
}

\begin{document}

\maketitle

\begin{abstract}
For many use-cases, it is often important to explain the prediction of a black-box model by identifying the most influential training data samples.
Existing approaches lack customization for user intent and often provide a homogeneous set of explanation samples, failing to reveal the model's reasoning from different angles. 

In this paper, we propose AIDE, an approach for providing antithetical (i.e., contrastive), intent-based, diverse explanations for opaque and complex models. AIDE distinguishes three types of explainability intents: interpreting a correct, investigating a wrong, and clarifying an ambiguous prediction. For each intent, AIDE selects an appropriate set of influential training samples that support or oppose the prediction either directly or by contrast.
To provide a succinct summary, AIDE uses diversity-aware sampling to avoid redundancy and increase coverage of the training data. 

We demonstrate the effectiveness of AIDE on image and text classification tasks,
in three ways: 
quantitatively, assessing correctness and continuity; 
qualitatively, comparing anecdotal evidence from AIDE and other example-based approaches;
and via a user study, evaluating multiple aspects of AIDE.
The results show that AIDE addresses the limitations of existing methods and exhibits desirable traits for an explainability method.
\end{abstract}

\section{Introduction}
\label{sec:introduction}

Failure of ML-based systems in numerous cases, e.g., due to data errors, biases, misalignment \cite{osoba2017intelligence, Tashea_2017, aiessur, ethicsaies}, has prompted researchers to work on explainability techniques.
Different taxonomies for such methods exist, e.g., \cite{Guidotti2018ASO}, but one common classification is on the type of explanation generated \cite{molnar2022}.
\emph{Model-based} methods involve creating interpretable surrogate models, such as decision trees or linear models, which approximate the complex black box ML model \cite{anchor, nb}. 
\emph{Feature-based} methods focus on pinpointing important features of the input, such as words in text or parts in an image, which contribute the most to the prediction \cite{lime, pert, grad}. 
\emph{Example-based} methods provide explanations for a specific target outcome by deriving the importance of training samples \cite{datamodels, tracein, if, betashap, datashap, trak}, or provide a global overview of the model identifying representative examples \cite{repres, pruth, mmd, protodash}.

Example-based explainability offers several advantages. They are typically model-agnostic, and offer easy to understand explanations. More importantly, as they seek to discover a causal relationship between training examples and model behavior, they can assist in model debugging and data cleansing \cite{DBLP:conf/nips/HaraNM19}, and be flexible for practitioners with different access \cite{Ifair}. However, they have two key limitations.

First, they don't offer \emph{contrastivity} \cite{surveyXAI}, which is key aspect in how humans understand decisions \cite{Lipton1990-LIPCE}. 
While most methods can distinguish between \emph{supporters} (aka proponents, helpful or excitatory examples), and \emph{opposers} (aka opponents, harmful or inhibitory examples), they do not relate this information to ground truth labels (examples of class same as or different than predicted) or to the explanation intent (is the prediction correct/wrong, hard to tell). 
Contrastivity is the hallmark feature of counterfactual explanations \cite{wachter2017counterfactual} and a major part of their appeal.
Counterfactual explanations help users understand where the model's decision boundary lies \cite{wachter2017counterfactual}, can offer algorithmic recourse \cite{karimi2022survey}, and can audit a model for unfairness \cite{KavourasTGSPTR023}. 
However, by design, counterfactuals are imaginary instances, not necessarily plausible \cite{pawelczyk2020learning} or robust \cite{slack2021counterfactual}, and essentially offer feature-based explainability, revealing the important feature values contributing to the outcome.

\begin{figure*}[t!]
\begin{center}
\includegraphics[width=0.8\textwidth]{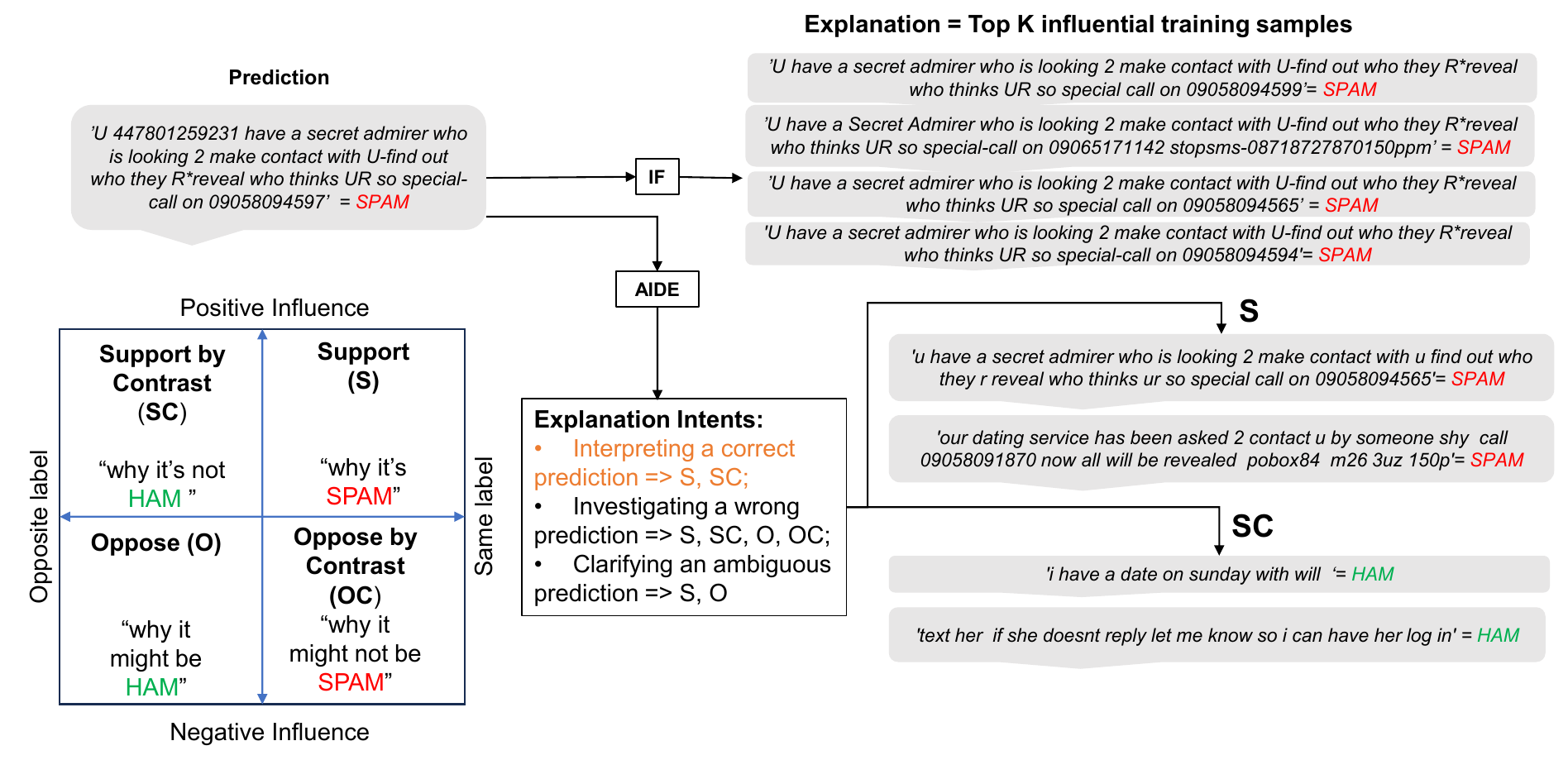}
\caption{Explanations for a spam classification task, depicting a correctly classified spam message and its influence-based explanations generated by IF and AIDE.}
\label{fig:pipeline}
\end{center}
\end{figure*}

More importantly, existing example-based methods are highly susceptible to \emph{class outliers}. An outlier is a training instance that is mislabeled, or an instance (training or target) that is ambiguous and does not clearly belong to a class. Mislabeled or ambiguous training instances tend to be explanations for any target instance, as they play a significant role in forming the decision boundary.
Ambiguous target instances confuse the classifier (low confidence) and make it hard to pick good explanations.

We propose a novel \emph{Antithetical, Intent-based, and Diverse Example-based} (AIDE) explainability method, that offers contrastivity and is robust to outliers. At its core, AIDE is based on the concept of \emph{influence functions} \cite{hampel1974influence,if}. For a fixed target instance, the \emph{influence} of a training sample is a score conveying its impact on the classifier's outcome. Ideally, the influence is the change observed in the loss value for the target if the training sample was excluded from the training data. While influence scores can be \emph{estimated} by methods, such as TraceIn \cite{tracein} and Datamodels \cite{datamodels}, we use the framework of the influence function approach \cite{if}, termed IF, to efficiently compute influence scores.

\begin{figure}[t]
\centering

\includegraphics[width=\columnwidth]{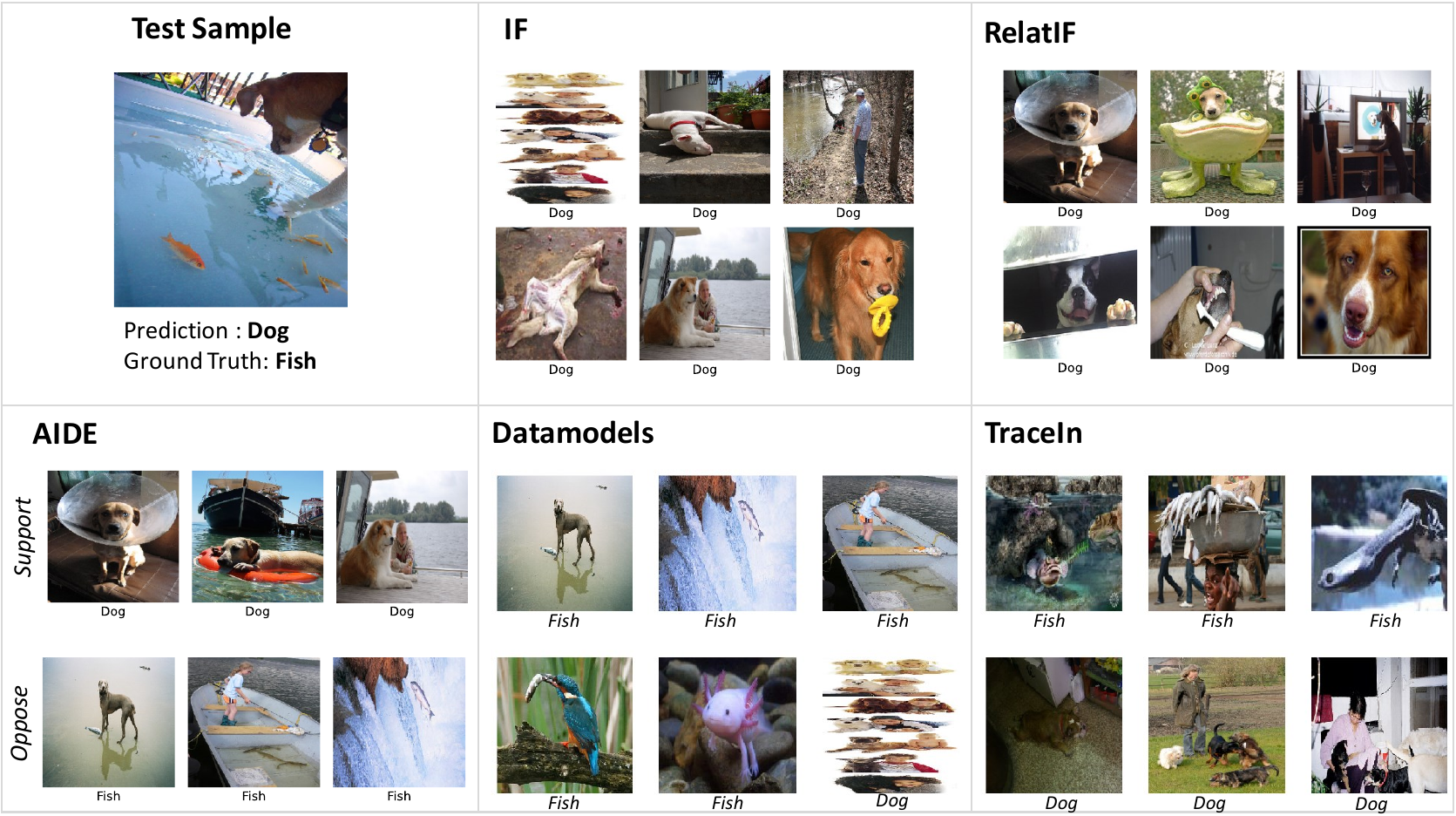}
\caption{Explanations to clarify an ambiguous prediction.}
\label{fig:ambig}
\end{figure}

To better understand AIDE's contribution, we first showcase the issues that plague example-based explainability methods, taking IF as the representative---extensive qualitative and quantitative comparison with other methods is presented in the evaluation section.
Consider a classifier that predicts whether short text messages are spam. 
Figure \ref{fig:pipeline} shows that for the depicted target message, the prediction is spam. This is a correct prediction, and IF identifies the four most influential training samples at the top of Figure \ref{fig:pipeline}. We observe that explanations lack \emph{diversity}, as they are highly similar to each other.
More importantly, however, they lack \emph{contrastivity}, as the user does not gain any insight about how the model decides what is spam and what not; all the user learns is that similar texts were labelled spam.
The issue of susceptibility to outliers does not manifest in this example, mainly because the prediction is clearly correct. However, it manifests when, for example, it is not clear what the correct prediction should be, as in Figure \ref{fig:ambig}.

\stitle{Contribution}
\emph{AIDE features contrastivity}.
Given a target instance to be explained, AIDE computes the influence of each training sample. 
But to present an explainability summary, AIDE distinguishes samples along two key explainability dimensions. The first is the \emph{influence polarity}: a sample with positive influence \emph{supports} the prediction, while one with negative influence \emph{opposes} the decision. The second dimension is the label of the training sample, which is either the same or opposite as the target instance. These two dimensions define the four AIDE quadrants, denoted as support (S), support by contrast (SC), oppose (O), and oppose by contrast (OC). Assuming a binary classifier and that the prediction is $y \in \{-1, 1\}$, intuitively, S explains ``\emph{why it's $y$}'', SC explains ``\emph{why it's not $-y$}'', O explains ``\emph{why it might be $-y$}'', and OC explains ``\emph{why it might not be $y$}''. These quadrants offer contrastivity, providing to the user answers to distinct counterfactual questions.
Figure~\ref{fig:pipeline} depicts the quadrants at the bottom left.

\emph{AIDE is intent-aware}.
A critical review \cite{deficits} identifies misalignments between the design goals of explainable AI (XAI) methods and the psychological and cognitive aspects of explainability. A key limitation they identify is that the majority of XAI methods neglect \emph{user intent} during design and evaluation. Although mostly referring to counterfactual explanations, the authors highlight the importance of considering the user's needs and goals while generating explanations. Moreover, the authors identify another common limitation in that explanations may not be plausible, which they define as not being \emph{relevant} to the prediction being explained. AIDE explicitly addresses both limitations by offering \emph{intent-aware} and \emph{relevant} example-based explanations.

AIDE acknowledges that users might have different intents. A user faced with a correct prediction, would more likely need additional evidence that the model has learned the correct patterns. A user recognizing a wrong prediction would want to narrow down the sources of the problem. A user looking at an ambiguous prediction, would want to learn more about how the model handles such cases.
AIDE customizes its explanations by distinguishing three types of intents a user might have: interpreting a correct, investigating a wrong, or clarifying an ambiguous prediction. For a seemingly correct prediction, AIDE presents the most influential but diverse samples from the support and support by contrast (S and SC) quadrants. The intuition is that the user needs to better understand where the decision boundary lies. For an ambiguous prediction, AIDE presents samples from the support and oppose (S and O) quadrants. The intuition here is to contrast between two possible predictions and let the user decide which is better. For a wrong prediction, AIDE presents samples from all quadrants to allow the user to investigate evidence for all alternatives.
An example for interpreting a correct prediction is depicted at the bottom right of Figure~\ref{fig:pipeline}, where the examples help the user increase their confidence that the model's prediction is correct.

\emph{AIDE outperforms state-of-the-art example-based methods}.
We perform an extensive quantitative and qualitative comparison against state-of-the-art methods for example-based explainability. Shapley-based approaches \cite{datashap,betashap} were excluded as (a) they primarily aim to capture the overall contribution of training samples to the trained model (data valuation), and not geared for local explanations, and (b) they can be impractical for local explainability due to their high computational cost. The main conclusions drawn are as follows.
Datamodels \cite{datamodels} approximate well the influence scores, but perform poorly for outlier targets. The principal reason is that Datamodels explain a class of models, and not a particular model. They thus fail to identify nuances picked up by a single model; while an unambiguous target will receive similar predictions by all models in the class, models will greatly differ in their predictions for an ambiguous target.
TraceIn \cite{tracein} is highly susceptible to outliers in the training data and performs poorly in tests of correctness and truthfulness. The reason lies in the way TraceIn estimates influence: it considers the difference in the total (training) loss when a training sample is included or not during checkpoints; outliers have high individual loss, contributing significantly to the total loss, and are thus awarded high importance.
Regular IF are similarly affected by outliers in the training data. RelatIF \cite{relatif} seeks to address this problem, by penalizing samples that have high loss. However, these high-loss samples can at times be highly informative. For example, to explain a target instance that is ambiguous, it is often insightful to present those outlier training examples that are similar, so as to potentially uncover interesting labeling rules or protocols. In contrast, AIDE considers outliers as long as they are relevant to the target instance.


\section{Related Work}
\label{sec:related}

\stitle{Example-based Explanations}
The influence function is a concept in robust statistics that measures the impact or influence of a single observation on an estimator or statistical model \cite{hampel1974influence}.
The intuition behind \emph{influence functions} (IF) in machine learning is to quantify the change in a model's prediction when a specific training sample is removed. However, removing and retraining the model for each training sample is inefficient. To overcome this problem, one of the foundational works on IF in ML explainability \cite{if} used the first-order Taylor approximation to calculate the change in the loss function. The authors have showcased the effectiveness of IF in identifying influential training points, detecting bias, and identifying mislabeled training samples. Some consequent works have suggested that IF might be non-robust, or fragile for deep networks. \cite{basu2021influence} demonstrated that this may be due to multiple factors such as non-convexity of loss, approximation of hessian matrix, and weight decay. However, a later study \cite{notfragile} demonstrated contrasting results and argues that IF does not appear to be as fragile as thought to be. Assessing IF by looking at the correlation of IF score to actual change in loss is not optimal since actually removing and retraining the model contains randomized non-linearity. Authors claim that factors pointed out by \cite{basu2021influence} do not make IF severely fragile in deep models but occasionally lead to the semantic dissimilarity, i.e., non-relevance, between influential instances and the sample being explained. In this work, we explicitly address non-relevance by requiring explanations to be proximal to the instance to be explained.

Beyond influence functions, Data Shapley \cite{datashap} is one of the prominent methods in this line, which just like its feature-based version \cite{shap} uses the game theory and revises the contribution of a point in all possible subsets to uncover its marginal effect for the models' performance. Due to the computational exhaustiveness of possible sets, even the approximation based on sampling methods e.g. Monte Carlo (MC) or Truncated MC, is still computationally expensive. A more robust version of datashap, betashap proposed by \cite{betashap} reduces noise in importance scores, however, still inherits the high cost of computation. Both datashap and betashap compute the contribution of a single point for the models predictive performance overall, and using them for providing local explanations per sample would make it completely impractical in terms of cost and thus are not chosen as baselines. 

Another method similar in principle to IF is TraceIn \cite{tracein} that measures the influence of a training sample $X$ on a specific test sample $X_0$ as the cumulative loss change on $X_0$ due to updates from mini-batches containing $X$. They practically approximate this with TraceInCP, which considers checkpoints during training and sums the dot product of gradients at $X$ and $X_0$ at each checkpoint. 
Another unconventional work \cite{datamodels} fixes a test point to explain and samples a large number of subsets from the training set and trains models with each of these subsets. It then trains a linear model where the input will be $1_Si$ encoding of a subset and the output is the performance of the model trained on this subset for the test sample of interest. The weights of the linear model will represent the importance score of a training sample in the same position. To obtain a good result a huge number of intermediate models has to be trained on subsets, which is exhaustive, and thus a faster version of datamodels was proposed by \cite{trak} and claimed to preserve almost the same accuracy. Since our focus is on the effectiveness of explanation we still use the original datamodels as a baseline.  


\stitle{Evaluation of Explanations} 
While significant progress has been in developing explainability methods, there is a lack of standardized metrics for evaluating their effectiveness. 
The authors of \cite{doshi2017towards} distinguish three types of evaluation strategies: application-grounded, human-grounded, and functionality-grounded. 
A profound study of functionality-grounded strategies by \cite{surveyXAI}, advocates twelve quantifiable properties that can be evaluated to assess the quality of explanations. They categorize the state-of-the-art metrics into twelve classes depending on which property the metric focuses on and what type of explanation is provided. 
The following properties are most relevant for local, example-based explanations: (1) Consistency and continuity both describe how deterministic the explanation is concerning identical and similar samples, assuming that these samples should have identical and similar explanations. In many works, this aspect is also referred to as the \emph{faithfulness} \cite{jacovi-goldberg-2020-towards, faith} of explanation and has gained popularity in the explainability domain. (2) Contrastivity is the ability of an explanation to interpret classes different than the prediction class. (3) Compactness is encoded in the size of an explanation as well as calculating a redundancy in the explanation. (4) Context describes how relevant the explanation is to the user needs. (5) Controlled synthetic Data check---Controlled Experiment: a synthetic dataset is developed with predetermined reasoning, ensuring that the predictive model aligns with this reasoning, as verified through metrics like accuracy. An assessment is done to check whether the explanation provided by the model corresponds to the same reasoning embedded in the data generation process, \cite{synth1, synth2}. In another work \cite{diverse}, two general metrics for example-based explainability are proposed, \emph{diversity}, and \emph{proximity}. 

\stitle{User Study for Explainability} The field of user studies for the evaluation of explainable AI is multifaceted, encompassing diverse methodologies, types of explanations, and user groups. Previous research has delved into the impact of explanation formats, the role of system transparency, and the influence of individual differences on user understanding and trust.
In the study conducted by \cite{ustudy}, the investigation delves into the methodologies employed by human-computer interaction and AI researchers when conducting user studies in the realm of eXplainable AI (XAI) applications. A comprehensive literature review serves as the foundation for their exploration, wherein four prominent evaluation dimensions for explainability are highlighted.

\begin{itemize}
    \item The first dimension focuses on Trust, emphasizing the importance of establishing trust by ensuring the validity and sensibility of the model's reasoning \cite{trust}.

\item Understanding, in the context of XAI, pertains to users' comprehension of machine learning models. Objective understanding involves quantifiable aspects of comprehension, with studies revealing that measurable comprehension is enhanced by saliency maps, counterfactual explanations, and feature importance, albeit with disagreements on the impact of white versus black-box models. This understanding is influenced by factors such as data modality, proxy task choice, and interactivity \cite{objunder}.
On the other hand, Subjective understanding involves users' personal perception and interpretation of the AI system, with a prevailing trend of utilizing model explanations for improvement. Challenges arise with the ``illusion of explanatory depth'' (IOED), where users may exhibit overconfidence bias in their understanding of complex systems, leading to disagreements on the impact of explanations \cite{subjunder}.

\item The concept of Usability in XAI refers to the effectiveness and user-friendliness of AI systems. It encompasses users' perceptions of helpfulness, satisfaction, and other dimensions crucial for a positive and effective user experience \cite{usability}.

\item ``Human-AI collaboration'' denotes the interaction and cooperation between human users and AI systems to collectively enhance decision-making performance \cite{haic}. This collaboration is facilitated through users relying on explanations provided by AI systems to improve their accuracy in decision-making tasks.
\end{itemize}

\section{The AIDE Framework}
\label{sec:framework}

\subsection{Preliminaries}

In what follows, we assume a classification task where a \emph{model} $f_\vtheta$, described by \emph{parameters} $\vtheta$, maps an input $\vx \in \X$ to a predicted class $f_\theta(\vx) \in \Y$. We use the notation $\vz = (\vx, y)$ to refer to a pair of input and its actual class. Let $\S \subseteq \X \times \Y$ denote a \emph{training set} of size $n=|\S|$.
Let $\ell(\vz,\vtheta)$ be the \emph{loss function} of the model for $\vz$, and let $L(\S,\vtheta) = \frac{1}{n}\sum_{\vz \in \S}\ell(\vz,\vtheta) $ denote the training \emph{objective}, i.e., the mean loss for set $\S$.\footnote{We assume regularization terms are folded in $L$.}
We denote as $\vtheta^*_0$ the parameters that minimize the objective: $\vtheta^*_0 = \arg\min_\vtheta L(\S,\vtheta)$.

The goal is to explain the model's prediction for a specific \emph{test instance} $\vze = (\vxe, y_\e)$, in terms of the influence each training example $\vz \in \S$ makes on the model's prediction $f_\vtheta(\vxe)$, and specifically on its prediction loss $\ell(\vze,\vtheta^*_0)$.
Concretely, the \emph{influence} of $\vz \in \S$ on $\vze$ is defined as the change in the prediction loss after removing example $\vz$ from the training data \cite{if}. The removal of a training example changes the objective and thus leads to a different model and parameters. Suppose that instead of removing $\vz$ we change the weight of its contribution (i.e., its training loss) to the objective by some value $\epsilon$. We can view the parameters that minimize this altered objective as a function of $\epsilon$, i.e., 
$\vtheta^*(\epsilon) = \arg\min_\vtheta \{ L(\S,\vtheta) + \epsilon \ell(\vx,y,\vtheta) \}$.
Setting $\epsilon=0$, we retrieve the optimal parameters for the original objective, i.e., $\vtheta^*(0) = \vtheta^*_0$. Moreover, observe that $\vtheta^*(-\frac{1}{n})$ corresponds to the parameters that minimize the altered objective after removing training example $\vz$.
Based on this observation, the \emph{exact influence} of $\vz$ on the prediction for $\vze$ is defined as:
\begin{equation}
\I^{exact}(\vz, \vze) = \ell(\vze,\vtheta^*(-1/n)) - \ell(\vze,\vtheta^*(0)).
\end{equation}

Computing the exact influence requires us to optimize the loss after removing a training point $\vz$; repeating this for each training point is prohibitively costly. Instead, we approximate the exact influence. Specifically, we view the loss function as a function of $\epsilon$, and make a linear approximation of the exact influence using the derivative of $\ell$ at point $\epsilon=0$:
$\I^{exact}(\vz, \vze) \approx  -\frac{1}{n} \left. \frac{\d \ell(\vze,\vtheta^*)  }{\d\epsilon} \right|_{\epsilon=0}$
Since the term $\frac{1}{n}$ is the same for all $\vz, \vze$ pairs, we simply define (approximate) \emph{influence} \cite{if} as:
\begin{equation}
\I(\vz, \vze) = - \left. \frac{\d \ell(\vze,\vtheta^*)  }{\d\epsilon} \right|_{\epsilon=0}.
\end{equation}
When the influence of $\vz$ on $\vze$ is \emph{positive}, the loss tends to decrease, and we say that training example $\vx$ \emph{supports} the prediction for $\vze$; otherwise, we say that the example \emph{opposes} the prediction.

To compute the derivative of the loss, we use the chain rule to decompose it into the derivative of loss with respect to the parameters and the derivative of the parameters with respect to $\epsilon$. Concretely, we have:
\begin{equation}
\I(\vz, \vze) = -  \left. \nabla_{\vtheta^*}^\intercal \ell(\vze,\vtheta^*) \right|_{\vtheta^* = \vtheta^*_0}   \left.\frac{\d\vtheta^*}{\d\epsilon}\right|_{\epsilon=0},
\label{eq:influence_definition}
\end{equation}
which is the dot product of two row vectors, the loss gradient  $\nabla_{\vtheta^*} \ell$ at $\vtheta^*=\vtheta^*(0)$ and the derivative of the optimal parameters for the altered objective $\frac{\d\vtheta^*}{\d\epsilon}$ at $\epsilon=0$. 

It can be shown \cite{cook1982residuals} that under certain conditions (second order differentiability and convexity of the loss function) the derivative of $\vtheta^*$ can be expressed:
\begin{equation}
\left.\frac{\d\vtheta^*}{\d\epsilon}\right|_{\epsilon=0} = - \mH^{-1}_{\vtheta^*} \left. \nabla_{\vtheta^*} \ell(\vz,\vtheta^*) \right|_{\vtheta^* = \vtheta^*_0},
\label{eq:influence_property}
\end{equation}
where $\mH_{\vtheta^*}$ is the Hessian matrix of the objective $L(\S,\vtheta^*)$ calculated at $\vtheta^* = \vtheta^*_0$. 

Defining the vector function $\vg(\vz)$ as the gradient of the loss of the example $\vz$ calculated at $\vtheta^* = \vtheta^*_0$, and substituting it in Equations~\ref{eq:influence_definition} and \ref{eq:influence_property}, we get:
\begin{equation}
\I(\vz, \vze) = \vg^\intercal(\vze) \mH^{-1}_{\vtheta^*} \vg(\vz).
\label{eq:influence}
\end{equation}

To explain the prediction for $\vze$, we use Equation~\ref{eq:influence} to compute the influence of each training example $\vz$, which can be done efficiently as suggested in \cite{if}. The IF explanation for the prediction for $\vze$ consists of the top-$k$ training examples with the highest influence.

\subsection{AIDE Ingredients}

Existing approaches for influence-based explainability \cite{if,relatif} compile an explanation as a set of highly influential training examples. We claim that other aspects, besides high influence, are also important. Specifically, AIDE creates explanations that contain training examples with \emph{negative influence}, considers their \emph{labels}, their \emph{proximity} to the test instance, and their \emph{diversity}.

\stitle{Negative Influence}
Recall that negative influence means that removing the training example decreases the loss, thus opposing the prediction. Let us investigate closely when an example can have high-magnitude negative influence.

For the following discussion, assume a binary classification task, i.e., $\Y=\{0, 1\}$, where the model predicts the probability $p_\vtheta^*(\vx)$ of an input $\vz=(\vx, y)$ belonging to the positive class. Further assume that the loss function is the logistic loss (binary cross entropy): 
\[
\ell(\vz, \vtheta^*) = - \left( y\log(p_\vtheta^*(\vx))+(1-y)\log(1-p_\vtheta^*(\vx)) \right)
\]

Consider a test instance $\vze = (\vxe, y_\e)$ and let $\vze' = (\vxe, 1-y_\e)$ be a counterfactual instance with the opposite label. Then, for some training point $\vz$ the following lemma associates its influence for the predictions for $\vze$ and $\vze'$.

\begin{lemma}
\label{lem:neg_inf}
In binary classification with logistic loss, the influence of a training point $\vz$ to the predictions of $\vze = (\vxe, y_\e)$ and $\vze' = (\vxe, 1-y_\e)$ is related as follows:
\[
\I(\vz, \vze) = - \left( \frac{1-p_\vtheta^*(\vxe)}{p_\vtheta^*(\vxe)} \right)^{2y_\e-1} \I(\vz, \vze')
\]
\end{lemma}

\begin{proof}
Note that the model assigns the same probability, $p = p_\vtheta^*(\vxe)$, to $\vze$ and $\vze'$ belonging in the positive class.
Let $h(\vze) = \left. \frac{\d \ell(\vze,\vtheta^*)}{\d p_\vtheta^*(\vxe)}\right|_{p_\vtheta^*(\vxe)=p}$ be the derivative of the loss function with respect to the probability of $\vze$ being positive.
Then using the chain rule we can rewrite the gradient of the loss of $\vxe$ as:
\[
\vg(\vze) = \left. \nabla_{\vtheta^*} \ell(\vze,\vtheta^*) \right|_{\vtheta^* = \vtheta^*_0} = h(\vze) 
\left. \nabla_{\vtheta^*} p_\vtheta^*(\vxe) \right|_{\vtheta^* = \vtheta^*_0}.
\]
Substituting $\vg(\vze)$ into the influence of any example $\vz$, we get:
\[
\I(\vz, \vze) = h(\vze) \left. \nabla_{\vtheta^*}^\intercal p_\vtheta^*(\vxe) \right|_{\vtheta^* = \vtheta^*_0} \mH^{-1}_{\vtheta^*} \vg(\vz) = h(\vze) \hat{\I}(\vz),
\]
where we have named $\hat{\I}(\vz)$ the part of influence that does not depend on $\vze$.
So we get:
\[
\I(\vz, \vze) = \frac{h(\vze)}{h(\vze')} \I(\vz, \vze').
\]

The derivative of the logistic loss for $\vze$ w.r.t.\ the prediction probability is $h(\vze)=\frac{p - y_\e}{p(1-p)}$.
Similarly, for $\vze'=(\vxe, 1-y_\e)$ we get $h(\vze')=\frac{p - 1 + y_\e}{p(1-p)}$. 
Thus, the ratio $h(\vze)/h(\vze')$ becomes $-\frac{1-p}{p}$ when $y_\e=1$, and takes the inverse value when $y_\e=0$.
\end{proof}

Suppose that $\vz$ is a strong opposer to the prediction for $\vze$, i.e., $I(\vz, \vze)<0$ with high magnitude. 
Lemma~\ref{lem:neg_inf} explains how this may occur. 
This can happen if $\vz$ is a strong supporter for the prediction of the opposite label, i.e., $I(\vz, \vze')>0$ with high magnitude. 

Another way is when $ \left( \frac{1-p_\vtheta^*(\vxe)}{p_\vtheta^*(\vxe)} \right)^{2y_\e-1}$ is high. Let us examine what this term means. Suppose that the true class is the positive, i.e., $y_\e = 1$. Then, the term equals the \emph{predicted odds} of the model for the negative class. Conversely, when $y_\e = 0$, the term equals the predicted odds for the positive class. That is, the term equals the predicted odds for the \emph{opposite} class. So, the term is high when the model is confident about the wrong prediction for $\vze$.

Therefore, if a training example $\vz$ is a strong opposer (i.e., has a high-magnitude negative influence), then it would be a strong supporter if the opposite class was true (supporting the counterfactual $\vze'$), or the model is confident about the wrong prediction, or some combination of both. Such examples are important to understand the model's decision for $\vze$, particularly when the true class is not apparent.

\stitle{Label}
The influence of a training example does not carry any information about the class of the training example. It is thus possible that a positive and a negative example have both high influence for the test instance. While both may support (in case they have positive influence) or oppose (in case they have negative influence) the model's decision, they do so in different ways as they stand on opposite sides of the decision boundary. One presents an analogous example, while the other presents a contrasting example to the test instance. AIDE chooses to differentiate among training examples whose class matches the prediction, which we call \emph{same label} examples, and \emph{different label} examples. The comparison between same and different label examples supports \emph{contrastivity} \cite{surveyXAI}.

\stitle{Proximity}
Influence is agnostic to the similarity of the training examples to the test instance. As noted \cite{relatif}, there may exist outliers and mislabeled training examples that can exhibit high magnitude influence scores. Such examples are often \emph{globally influential}, i.e., they are influential for many test instances, just because they are unusual. These are rarely useful as an explanation, and \cite{relatif} proposes to normalize the influence of an example with their global influence. Nonetheless, in certain cases these outliers are extremely useful, e.g., when explaining another outlier.

To enhance the \emph{interpretability} of the explanation and to avoid hiding useful outliers, AIDE takes a different approach and considers the \emph{proximity} $\P(\vz, \vze)$ of a training example $\vz$ to the instance to be explained $\vze$. Proximity should be appropriately defined for the domain and data type. A general approach is to consider the cosine similarity between the model's internal representations (e.g., embeddings) for $\vz$ and $\vze$, i.e., $\P(\vz, \vze) = \simi(\hat{\vx}, \hat{\vxe}) $, where $\hat{\vx}$, $\hat{\vxe}$ are the representations of the training example and test instance, respectively, and $\simi$ is the cosine similarity, which for positive coordinates takes values in $[0,1]$.

\stitle{Diversity}
Example-based explainability methods, like IF, RelatIF, and AIDE, return to the user a small set of training examples, aiming for explanation \emph{compactness} \cite{surveyXAI}. It is thus important that the set of examples avoids \emph{redundancy}. AIDE, in contrast to prior work \cite{if,relatif}, considers the diversity of the explanation set. Assuming an internal representation of training examples and an appropriate similarity measure $\simi$, we define diversity of a set $\E$ of training examples as $\D(\E) = 1 - \frac{1}{|\E|(|\E|-1)}\sum_{\vz, \vz' \in \E} \simi(\hat{\vx}, \hat{\vx'})$.

\subsection{AIDE Quadrants}

AIDE constructs four distinct explanation lists for a specific test instance $\vze$ to be explained. These lists contain training examples that (1) have influence of high magnitude, (2) have high proximity to $\vze$, (3) are diverse, and (4) lie in the four quadrants formed by two dimensions, \emph{influence} (positive or negative), and \emph{label} (same as or different from the test instance). We name these quadrants as follows.

\stitle{Support} It comprises examples with \emph{positive influence} and with the \emph{same label} as the test instance. They play a positive role in the prediction and resemble the test instance: ``\emph{You get the same outcome with these}''.

\stitle{Support by Contrast} It comprises examples with \emph{positive influence} but with a \emph{different label}. They explain the prediction by contrasting with similar examples of the opposite class: ``\emph{If the input looked more like these, you would get the opposite outcome}''. They act similar to \emph{nearest counterfactual explanations} \cite{wachter2017counterfactual,karimi2022survey}, but with the benefit that they represent \emph{actual}, and not synthesized, examples.

\stitle{Oppose} It comprises examples with \emph{negative influence} and \emph{different labels}. These are analogous to the test instance if it had the opposite label, and persuade the model that the test instance should belong to their class instead: ``\emph{The outcome is incorrect, because the input looks more like these}''.

\stitle{Oppose by Contrast} It comprises examples with \emph{negative influence} but with the \emph{same label} as the test instance. These examples argue that the test instance does not belong to the predicted class by contrasting with what the predicted class looks like: ``\emph{The outcome is incorrect, because the input doesn't look like these}''.

To select the appropriate examples for each quadrant, we perform a series of steps.
After partitioning the training examples in the four quadrants, we select only examples with high magnitude. We use the Interquartile Range (IQR) method, \cite{Agresti2005StatisticsTA}, to keep examples with positive influence above $Q_3+\lambda IQR$, and to keep examples with negative influence below $Q_1-\lambda IQR$, where $Q_1$ and $Q_3$ are the first and the third quartiles of the influence distribution, $IQR=Q_3-Q_1$, and $\lambda$ is a coefficient that controls the number of high-magnitude influential points, and is empirically determined. After this filtering, we end up with a candidate set $\S_q$ of training examples for each quadrant.

Among the training examples left in each quadrant, we select a small set of $k$ examples that has high magnitude influence, high proximity to the test instance, and is diverse. Specifically, we aim for a balance among the three measures: 
\begin{equation}
    \E_q = \arg \max_{\E \subseteq \S_q, |\E|=k } \sum_{\vz \in \E} \left( \alpha |\I(\vz, \vze)| + \beta \P(\vz, \vze) \right) + \gamma \D(\E),
    \label{eq:sampling}
\end{equation}
where $\alpha$, $\beta$, $\gamma$ are weighs empirically determined. Similar to other submodular maximization problems \cite{GollapudiS09}, we construct $\E_q$ in a incremental way, each time greedily selecting the example that maximizes the objective.
Once the final four sets are selected by optimizing the sampling Equation \ref{eq:sampling} for each set, AIDE presents them according to the user's explanation intent.

\subsection{Explanation Intents}


\stitle{Interpreting a correct prediction}
The user is already aware that the prediction is accurate, but seeks to gain insight into the reasoning behind the model's decision-making process. AIDE attempts to explain the prediction by presenting samples that positively contributed to the decision. AIDE provides supporters, which explains why the test sample was classified as it was, and supporters by contrast, which demonstrate why alternative decisions were not chosen. Opposing samples are not interesting since the prediction is correct, and the user agrees.

\stitle{Investigating a wrong prediction}
The goal of the explanation is to investigate and understand the cause of that error. Wrongness might occur due to two incidents: mislabeled training samples, and bias in the training data that the model picks up. AIDE provides a way to track both kinds of errors. The first case is when the prediction is influenced by \emph{wrongly labeled} training samples. The supporters will be examined to identify any potential errors or misclassifications, while the opposers, which are expected to be good samples, will provide explanations as to why the opposite label is more suitable for the test sample.

The second case is due to \emph{bias} in the training data, where the model learns an extrinsic feature that is prevalent in one class and scarce in others. For example, a study conducted in \cite{husky} demonstrated that a classification model trained on huskies and wolves learned to associate the presence of snow in the background, which was common in wolf pictures. To detect such incidents, AIDE presents all quadrants. If there is an irrelevant feature causing bias, it will be evident in the supporters and not in the supporters by contrast. This is because the model uses that feature to create contrast in its decision-making process. Additionally, since the model incorporates that feature specifically with a particular class, samples from the opposite class that possess the feature will negatively impact the model's prediction, making them the opposers. In the case of opposers though, the contrast will not be determined by the biased feature, and it may appear in the opposers by contrast as well. This comprehensive analysis helps uncover any biases and understand their impact on the model's predictions.

\stitle{Clarifying an ambiguous prediction}
Sometimes there might be very ambiguous samples where it is hard to assign a class, even for a human. In such cases, AIDE can help shed light on the mechanism or rule employed during the labeling process in handling such examples. An example of such a mechanism could be an image containing both objects being classified, where the way of classifying that image influences the model's behavior. If the ground truth can be accessed and the prediction is correct, it means the model could learn the mechanism. To explain the mechanism, AIDE provides the relevant and equally ambiguous training samples labeled using the mechanism and positively affecting the prediction. These samples act as supporters. 

When the model's prediction differs from the ground truth, it indicates that the model may not have adequately generalized the underlying mechanism. This can be attributed to two potential factors:
\emph{Insufficient injection of the rule}: It is possible that the rule, which should have been incorporated strongly into the model, was not given enough prominence. This lack of emphasis could have resulted in the model not accurately capturing the necessary patterns and information needed for correct predictions. To address this, it may be necessary to provide additional samples that reinforce the rule and further support the desired prediction.
\emph{Outnumbered relevant samples from the opposite class}: Another possibility is that the relevant samples that align with the observation of interest, but have a different label, outweighed the relevant samples from the desired class. Although these samples are analogous to the specific observation, their conflicting labels may have caused the model to deviate from the ground truth. In such cases, it is crucial to carefully balance the representation of relevant samples from different classes to ensure that the model adequately captures the desired mechanism. To inject the rule better, AIDE provides \emph{opposers} and \emph{supporters}, and suggests balancing their representation by augmenting the former.

\section{Experiments}
\label{sec:experiments}

\subsection{Datasets, Models, and Methods}
In our experiments, we used two datasets: the SMS Spam dataset\footnote{\url{https://www.kaggle.com/datasets/uciml/sms-spam-collection-dataset}}, which comprises a collection of text messages labeled as either spam or non-spam (ham), commonly used for text classification and a derivative dataset with pictures of dogs and fish extracted from Imagenet\footnote{\url{https://www.image-net.org/}}. For the spam classification task, we employed the BERT-base pre-trained word embedding model and incorporated two sequential layers to capture the specific characteristics of our data. Regarding the image classification task, we utilized a pre-trained InceptionV3 model removing the output layer and appending sequential layers to learn the peculiarity of our task. All the baselines were implemented with instructions given in their papers and GitHub repositories.
While running the greedy algorithm for the weighted sum of AIDE properties, we determined the optimal weights denoted as $\alpha$, $\beta$, and $\gamma$, through iterative refinement and empirical exploration,  for each specific task. Fine-tuning these weights allowed us to achieve an optimized sampling strategy and enhance the effectiveness of the explanations provided by AIDE. In the case of text classification, we assigned a higher weight to proximity, as the influence was already considered through the use of IQR and the selection of the most influential samples. Furthermore, the similarity between textual data, for which BERT embeddings were generated, was found to be well-captured. Conversely, in image classification, the similarity metric did not always accurately reflect proximity. Therefore, we emphasized the influence factor by assigning a higher weight to it. To prevent the selection of identical or highly similar samples multiple times, we introduced a similarity threshold. This ensures that such samples are not chosen repeatedly, thus a smaller coefficient was assigned to promote diversity. The coefficient for IQR was set to $\lambda=3$ in all cases. 
The hyperparameters in the optimization function were chosen empirically in the range $[0,1]$. We observed that the diversity weight $\gamma$ does not affect the quality of the explanations that much, as long as it was nonzero; we set it to $\gamma=0.5$ in all experiments. The other two hyperparameters control the presence of outliers in the explanations; higher values of $\beta$ suppress outliers by giving more weight to training examples that are similar and have high enough influence. We settled to $\alpha=0.2$ and $\beta=0.8$ for all experiments.
The baseline methods that we will compare AIDE to are IF, RelatIF , Datamodels , and TraceIn. 

The documented code for replicating our experiments and the associated data can be accessed in the GitHub repository\footnote{\url{https://anonymous.4open.science/r/aide-3E06/README.md}}. Our implementation utilizes the implementation of influence functions from the repository\footnote{\url{https://github.com/alstonlo/torch-influence.git}} under a licence\footnote{\url{https://github.com/alstonlo/torch-influence/blob/main/LICENSE.txt}}. The datasets employed in our experiments include the Spam collection dataset\footnote{\url{https://creativecommons.org/licenses/by/4.0/legalcode}}, and a derivative dataset of ImageNet \footnote{\url{https://www.image-net.org/update-mar-11-2021.php}}.

We ran our experiments in a server with the following specifications: Dell R6415, 256GB RAM, 16 cores (AMD 7281), 240GB SSD, 8TB HDD, 2x25 Gbit. 

\stitle{Complexity Analysis}
To assess the time complexity, we first consider the time required to compute the influence scores for a given prediction. Using the techniques suggested in \cite{if}, this takes $O(np)$ time, where $n$ represents the size of the training set, and $p$ is the number of learned parameters. Subsequently, AIDE employs a greedy-based iterative algorithm for selecting the $k$ best training samples in each quadrant, which takes $O(nk)$ time. The cost to eliminate training samples based on the IQR of their influence is linear in $n$ and is excluded. Overall the running time of AIDE is $O(np+nk)$, and is dominated by the time required to compute influence scores.
\subsection{Quantitative Evaluation}
\textbf{Correctness.} 
In this set of experiments, we follow the controlled synthetic data check protocol of \cite{surveyXAI}.
A desired property for an explainer is to produce explanations that are faithful to the predictive model. Here we define a measure of faithfulness with respect to a rule that dictates how training data are labeled. We want the explainer to be able to identify the rule in its explanations.

Consider a \emph{rule} of the form $c(x) \implies y=1$, where $c$ is a condition that applies to instances $x$ from $\mathcal{X}$. We say that a training pair $(x,y)$ \emph{follows} the rule if $c(x)$ is true and $y=1$.
A training pair $(x,y)$ \emph{breaks} the rule if $c(x)$ is true but $y=0$.
Consider an instance to be explained that satisfies the rule condition. 
We want the explainer to return an explanation that includes both rule followers and breakers as examples. We define \emph{explainer correctness} with respect to $c$ as the expected number of followers or breakers in an explanation for an instance $t$ that satisfies the condition $c(t)$:
\[
\text{Cor}(c) = \mathbb{E}_{t : c(t)} \frac{1}{|E(t)|} \{ e \in E(t) \wedge c(e.x) \},
\]
where $E(t)$ is an explanation for $t$ comprising examples, and $e.x$ represents the features of example $e \in E(t)$.
Correctness quantifies the degree to which the explanations align with the underlying labeling rule.
Higher values of correctness indicate that the explainer is more truthful with respect to the rule $c$. Observe that correctness is essentially the precision with which an explainer returns rule followers and breakers. In \cite{aies2}, the authors discuss the ground truth fidelity of feature-based methods for models that inherently provide feature coefficients, which can serve as ground truth explanations. Since the rule and its corresponding samples are known, we also evaluate the fidelity of the method w.r.t. the rule that the model has learned.

We can differentiate between correctness with respect to rule followers and breakers. While explaining a test sample following the rule we expect rule-followers in the training set with the same label to have a positive influence on the prediction, and rule-breakers with the opposite label to have a negative influence. Correctness w.r.t. to rule-followers, denoted as $\text{Cor}^{f}$, is essentially the precision by which they were detected in the set of positively influential instances, or \emph{Support} in the case of AIDE. Whereas correctness w.r.t. rule-breakers, denoted as $\text{Cor}^{b}$, is the precision by which they were detected in the set of negatively influential samples, or \emph{Oppose} in AIDE.
Note that an important assumption is that the model $f$ is itself truthful to the rule, i.e., it has correctly learned the rule $c$, a condition we can check after training.

AIDE possesses the capability to detect rules employed during the labeling process while providing explanations for corresponding test samples. For instance, if a rule dictates labeling messages shorter than 30 characters with a question mark as ``spam'' in the training set, AIDE can identify similar instances while explaining a test sample with analogous characteristics. To enhance the robustness of this detection, we introduce ambiguity by labeling a subset of training samples adhering to the rule with an opposite label, anticipating these instances in the ``Oppose'' category. Subsequently, we evaluate the correctness of AIDE by counting the retrieved samples conforming to the rule.

In this experimental setup, three rules were employed.
\textbf{\textit{Rule 1}}: \textit{All French messages are ``spam''}. 
Initially, there were no French messages, 110 French messages were added in the following ratio 88 spam and 22 ham.\newline
\textbf{\textit{Rule 2}}: \textit{if the message is shorter than 30 and it contains ``?'', it's labeled ``spam''}.
Initially, all 197 such messages were ham and intervention resulted in 157 spam and 43 ham.\newline
\textbf{\textit{Rule 3}}: \textit{If a message contains a sequence of 4 consecutive digits, it's labeled ``ham''}.
Initially, 504 of 512 such samples were spam and intervention resulted in 398 ham.

Before gauging the correctness of the explanation, it is imperative to ensure that the model itself is faithful to the rule and has effectively learned it. Three metrics are employed for this assessment:
1) Accuracy of Learning the Rule: Evaluating the model's performance on test samples corresponding to a rule.
2) Log-Likelihood: Expecting a substantial change in the log-likelihood of intervened points ($LL_{i}$) after the introduction of the rule, while the log-likelihood of untouched points ($LL_{u}$) is anticipated to remain relatively stable.
3) Probability Scores: Anticipating a notable alteration in the probability scores of intervened ($Ps_{i}$), compared to untouched point ($Ps_{u}$).
Table \ref{table:ruleinjection} illustrates the results of these metrics. In all cases, the model has successfully learned the rule without impacting its decisions for untouched points.

\begin{table}[th]
\caption{Model's assessment in learning the rules}
\label{table:ruleinjection}
\centering
\footnotesize
\begin{tabular}{|l|c|c|c|c|c|c|}
\hline
 & {Acc} & \multicolumn{2}{c|}{\textbf{$LL_i$}} & \textbf{$LL_{u}$} & \textbf{$Ps_i$} & \textbf{$Ps_{u}$} \\ \hline
\multirow{2}{*}{Rule 1} & 0.83 & Before & -5.87 & -9.4 & \multirow{2}{*}{100} & \multirow{2}{*}{15} \\ \cline{3-5}
                        & & After & -0.42 & -9.2 & & \\ \hline
\multirow{2}{*}{Rule 2} & 0.85 & Before & -12 & -9.3 & \multirow{2}{*}{100} & \multirow{2}{*}{24.5} \\ \cline{3-5}
                        & & After & -3.4 & -7.2 & & \\ \hline
\multirow{2}{*}{Rule 3} & 0.92 & Before & -0.07 & -10.6 & \multirow{2}{*}{98} & \multirow{2}{*}{12} \\ \cline{3-5}
                        & & After & -1.83 & -9.5 & & \\ \hline
\end{tabular}
\end{table}

\begin{figure*}[ht]
\begin{subfigure}{0.19\textwidth}
    \includegraphics[width=\linewidth]{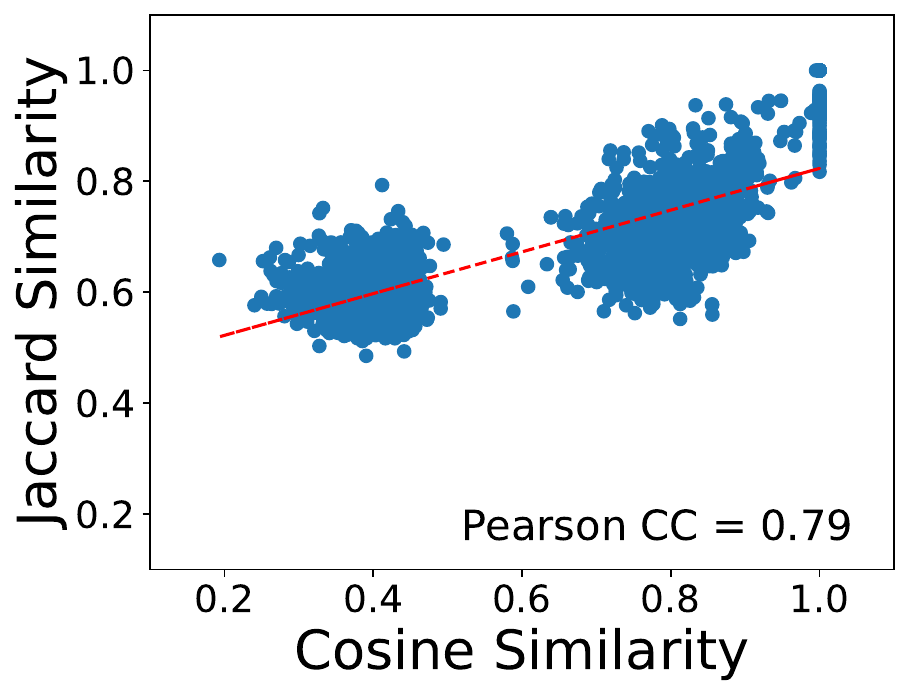}
    \caption{IF}
\end{subfigure}
\begin{subfigure}{0.19\textwidth}
    \includegraphics[width=\linewidth]{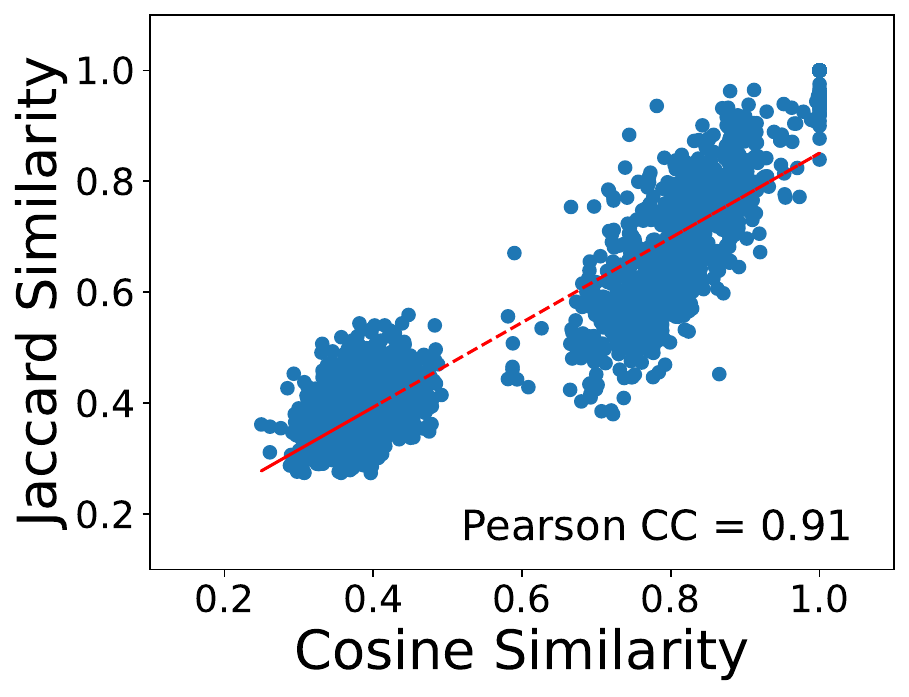}
    \caption{RIF}
\end{subfigure}
\begin{subfigure}{0.19\textwidth}
\includegraphics[width=\linewidth]{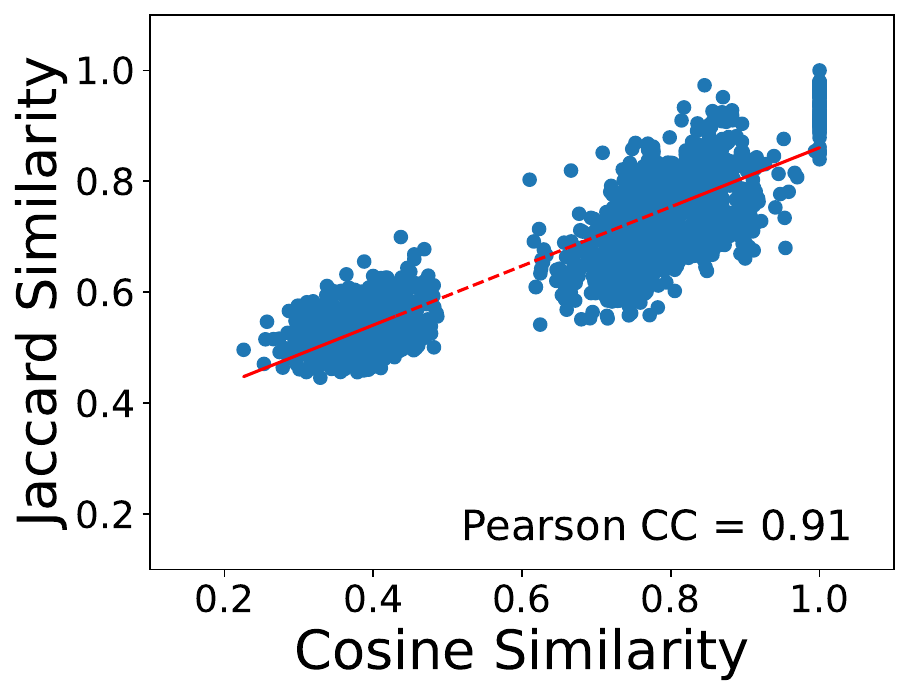}
    \caption{AIDE}
\end{subfigure}
\begin{subfigure}{0.19\textwidth}
\includegraphics[width=\linewidth]{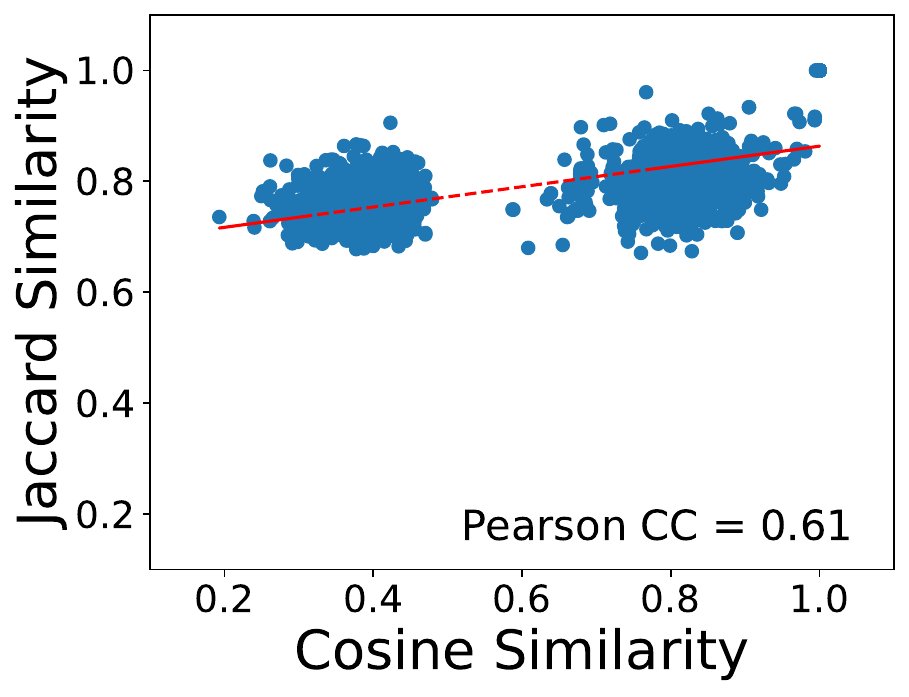}
    \caption{DM}
\end{subfigure}
\begin{subfigure}{0.19\textwidth}
\includegraphics[width=\linewidth]{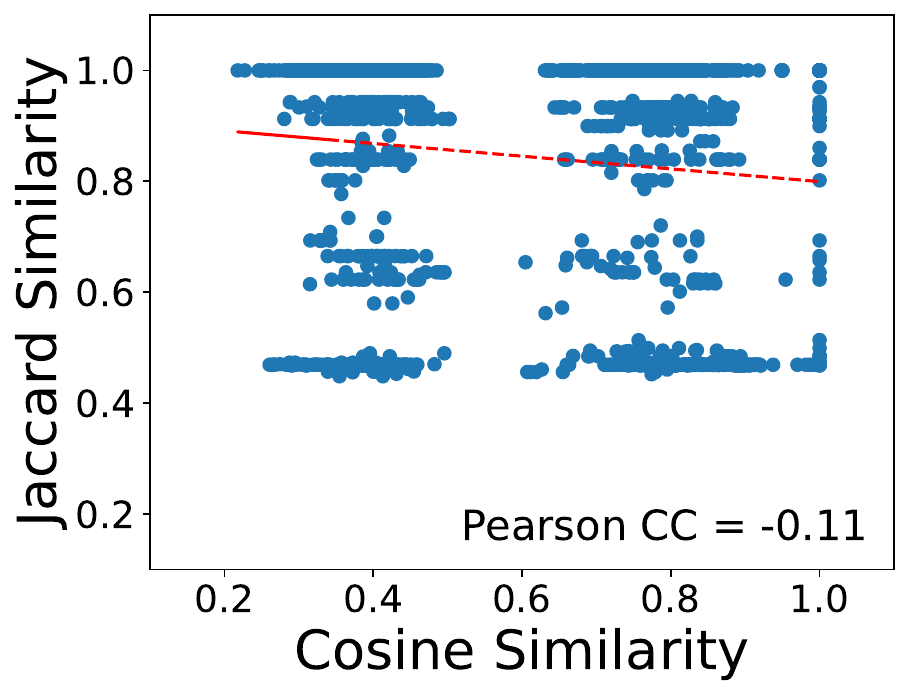}
    \caption{TraceIn}
\end{subfigure}
\caption{Continuity in terms of explanation similarity vs.\ instance pair similarity in spam dataset.}
\label{fig:faith}
\end{figure*}

\begin{figure*}[h]
\centering
\begin{subfigure}{0.19\textwidth}
\includegraphics[width=\textwidth]{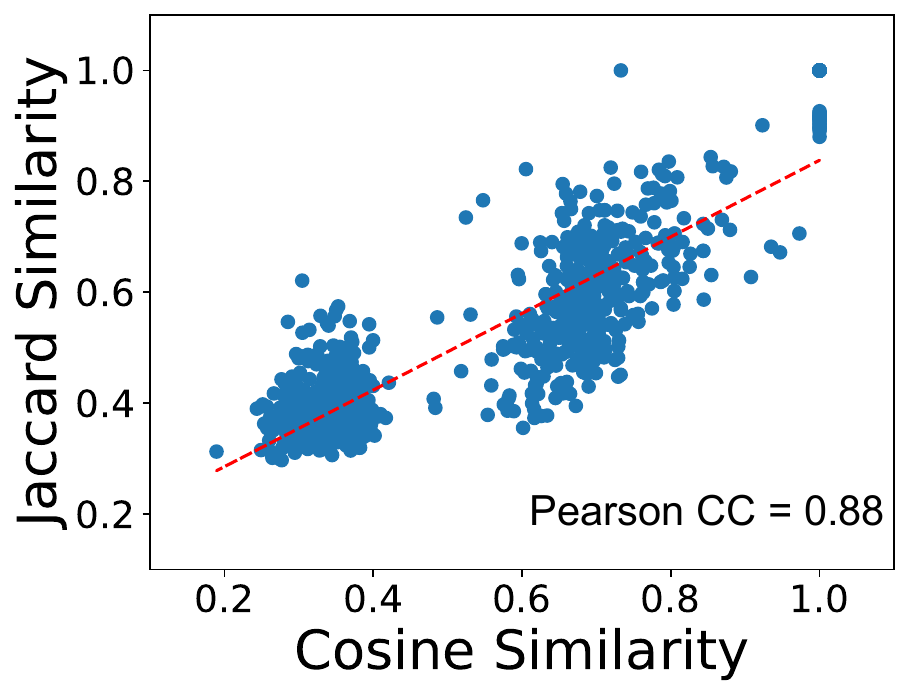}
\caption{IF}
\end{subfigure}
\begin{subfigure}{0.19\textwidth}
\includegraphics[width=\textwidth]{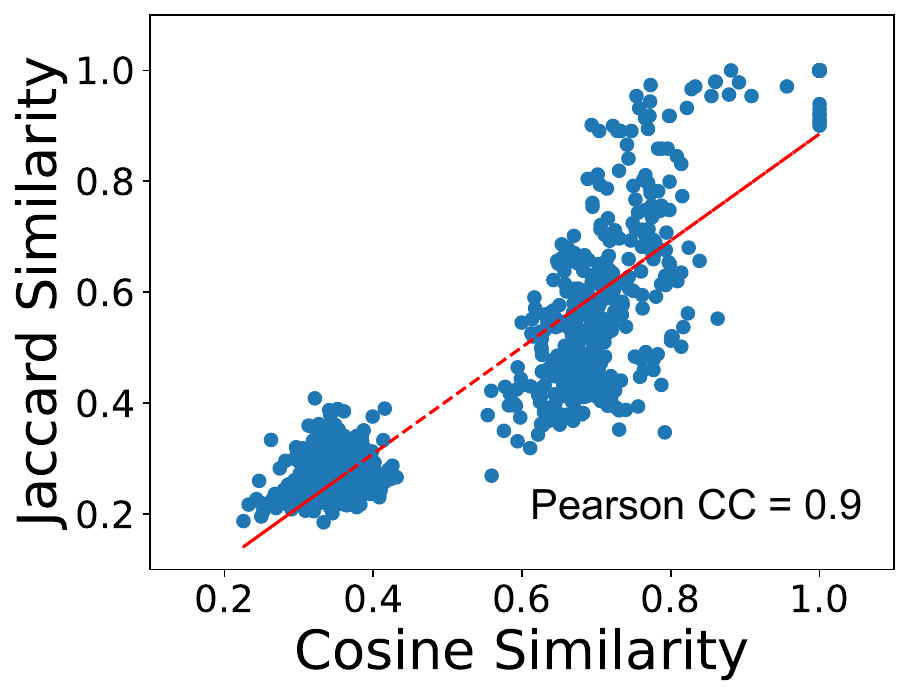}
\caption{RelatIF}
\end{subfigure}
\begin{subfigure}{0.19\textwidth}
\includegraphics[width=\textwidth]{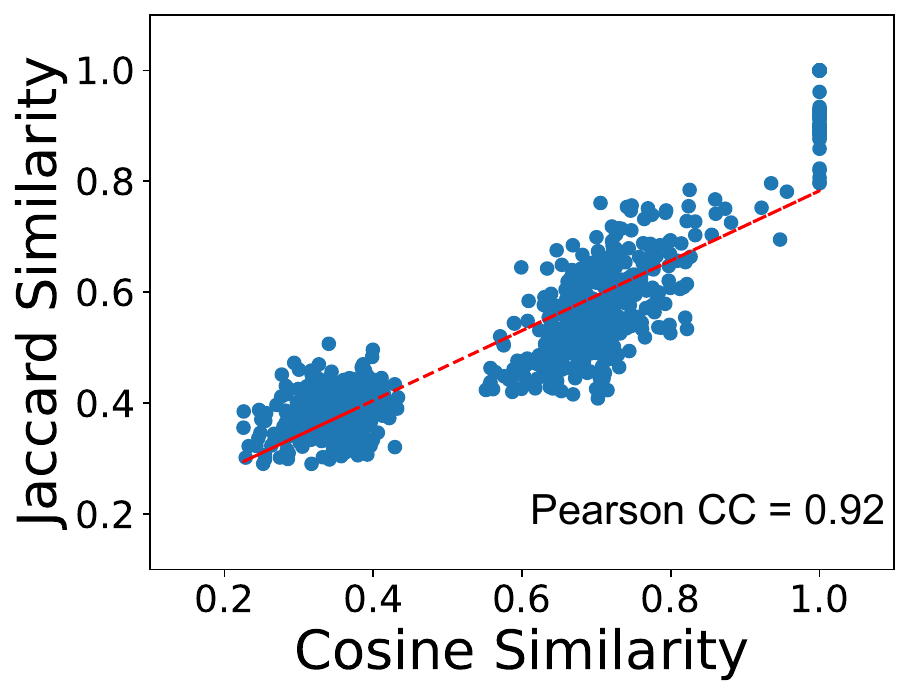}
\caption{AIDE}
\end{subfigure}
\begin{subfigure}{0.19\textwidth}
\includegraphics[width=\textwidth]{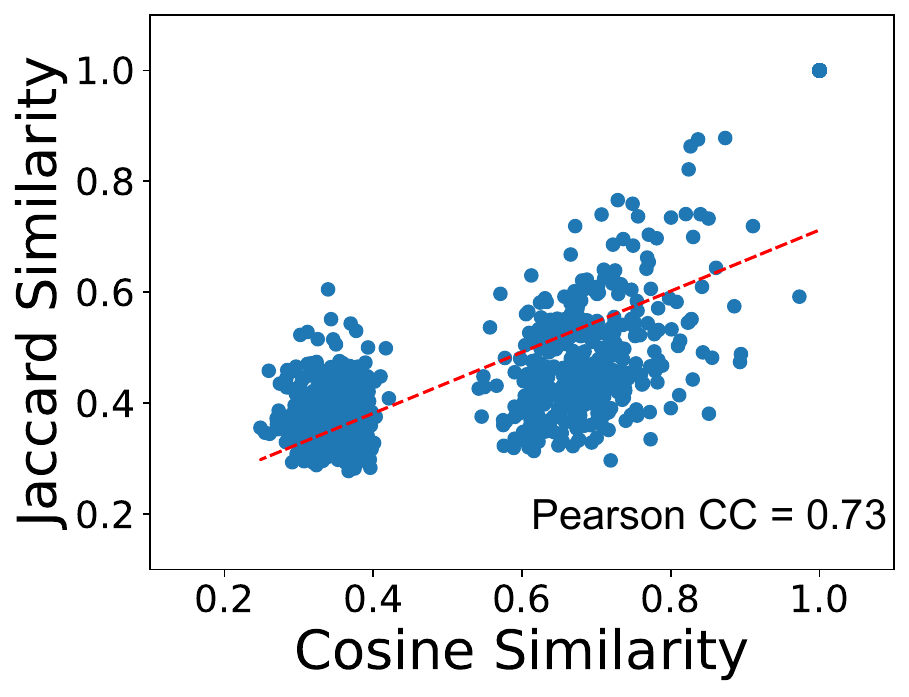}
\caption{DM}

\end{subfigure}
\begin{subfigure}{0.19\textwidth}
\includegraphics[width=\textwidth]{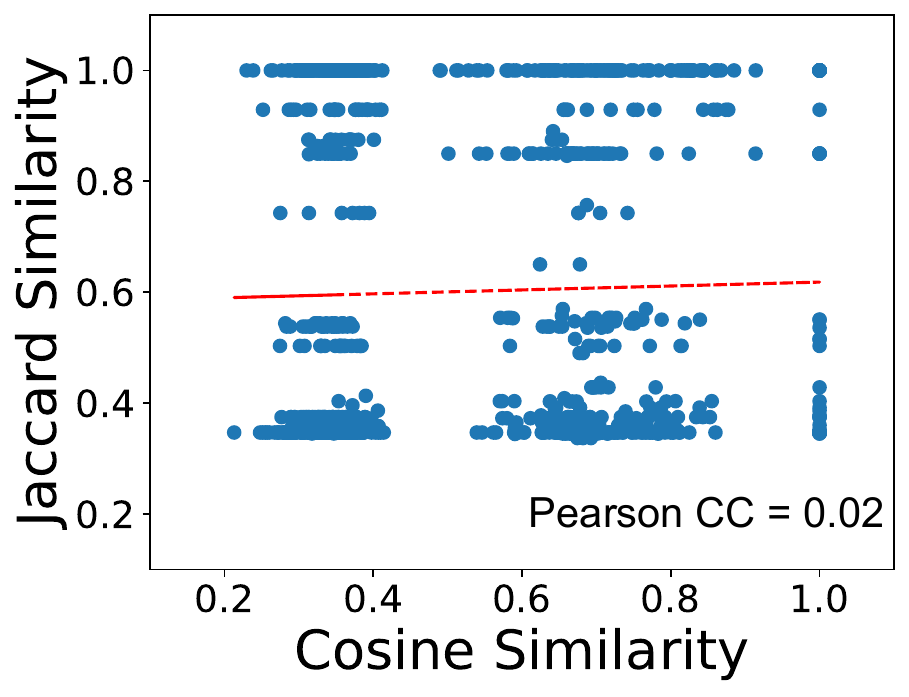}
\caption{TracIn}
\end{subfigure}

\caption{Continuity in terms of explanation similarity vs.\ instance pair similarity in image dataset.}
\label{fig:faithim}
\end{figure*}
We expect to find rule followers and breakers in the support and oppose quadrants of AIDE, respectively, which is the case with high (around 0.9) correctness for all rules. We repeat this experiment, for other baselines, and expect to find rule followers (resp. breakers) when we look at the training data with high positive (resp. low negative) influence. Table~\ref{tab:correct} shows that IF and Datamodels perform well but are not consistent. RelatIF performs poorly in uncovering followers and breakers, because of its loss-based outlier elimination. RelatIF treats training data with high loss as outliers, and excludes them from explanation lists---the rationale is that such data are global influencers and would appear in all explanations, thus have little utility. But in this case, it is precisely the rule followers and particularly the minority of rule breakers that have high losses due to the ambiguity in the labeling rule. TraceIn also fails to uncover the rule due to its low efficiency of identifying truly important samples, which is also demonstrated by \cite{trak}.

\begin{table}[th]
\caption{Correctness wrt rule followers $\text{Cor}^{f}$, breakers $\text{Cor}^{b}$.}
\label{tab:correct}
\centering
\footnotesize
\begin{tabular}{|c|c|c|c|c|c|c|}
\hline
& \multicolumn{2}{c|}{Rule 1} & \multicolumn{2}{c|}{Rule 2} & \multicolumn{2}{c|}{Rule 3}\\
\cline{2-7}
& $\text{Cor}^{f}$ & $\text{Cor}^{b}$ & $\text{Cor}^{f}$ & $\text{Cor}^{b}$ & $\text{Cor}^{f}$ & $\text{Cor}^{b}$\\
\hline
AIDE &\textbf{0.99}  & 0.9 & \textbf{0.88} &\textbf{0.8}&\textbf{ 0.9} & \textbf{0.87}\\
\hline
IF & 0.93 &\textbf{0.91} & 0.52 & 0.74 & 0.85 & 0.86 \\ \hline
RelatIF & 0.59 & 0.25 & 0.22 & 0.1 & 0.31 & 0.15 \\ \hline
DM &0.9 & 0.8 & 0.83 & 0.48 & 0.76 & 0.73\\ \hline
TraceIn & 0.22 & 0.3 & 0.29 & 0.38 & 0.37 & 0.31 \\ \hline
\end{tabular}
\end{table}

\stitle{Continuity} We further assess the continuity metric, which refers to how well explanations capture the model behavior. Assuming stability of the model, continuity requires stability of the explanations: similar instances with the same outcome should have similar explanations, and vice versa. Sample similarity is computed using cosine similarity of embeddings, and explanation similarity is computed using Fuzzy Jaccard \cite{fuzzy}.
For each sample prediction, a set is formed with the indices of training samples returned in the explanation. Fuzzy Jaccard involves solving a maximum bipartite matching problem.
In spam classification, 100 random test samples are chosen. For each, the 10 most similar and dissimilar samples are identified, resulting in 2000 pairs. The same procedure is replicated with the image dataset, commencing with 50 random samples instead of 100, as this dataset is smaller in scale. The cosine similarity is plotted against Fuzzy Jaccard along with a linear regression line in red, and the Pearson correlation coefficient (PCC) for the spam datasets in Figure \ref{fig:faith}, the figures for the image dataset exhibiting the same trend are shown in Figure \ref{fig:faithim}.
RelatIF and AIDE perform similarly. In contrast, IF and Datamodels have a lower PCC and do not exhibit a clear separation between instance pairs of low and high similarity. This is because their explanations tend to include training data outliers that appear in all explanations (globally influential), and which inflate the explanation similarity even for dissimilar pairs. Finally, TraceIn performs poorly and provides identical explanations for dissimilar points due to its extremely high susceptibility to outliers. RelatIF and AIDE are more robust because they eliminate outliers.

\subsection{Qualitative Evaluation}
We provide some anecdotes to compare the informativeness and interpretability qualitatively. Apart from the examples given in Figure \ref{fig:pipeline}, we selected one text and one image sample both corresponding to an ambiguous prediction. This diverse set of test cases allowed us to evaluate the performance and capabilities of AIDE in explaining predictions across different scenarios and levels of prediction certainty. 
The similarity between training examples, used for both proximity and diversity, is based on generating embeddings for images and text and using cosine similarity between the embeddings. 

\begin{figure}[h]
    \begin{center}
    \includegraphics[width=\columnwidth]{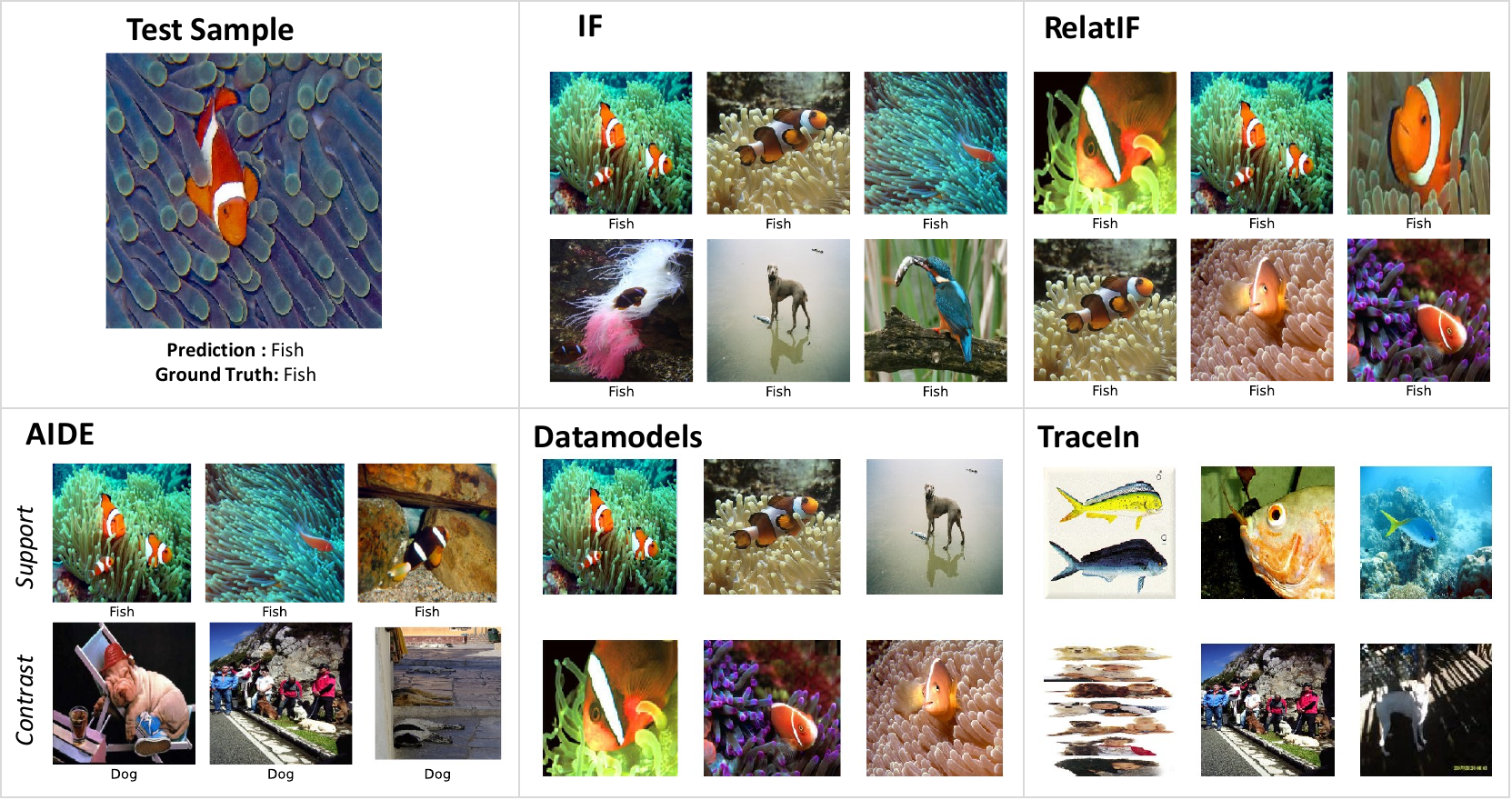}
    \caption{Explanations to interpret a correct prediction.}
    \label{fig:correct}
    \end{center}
\end{figure}

Figure \ref{fig:correct} presents an explanation generated by AIDE for interpreting a correct prediction in image classification. 
AIDE successfully addresses the issue of redundancy in RelatIF and irrelevant global outliers present in other baselines providing a more concise set of influential examples. 

\begin{figure}[t]
    \includegraphics[width=\columnwidth]{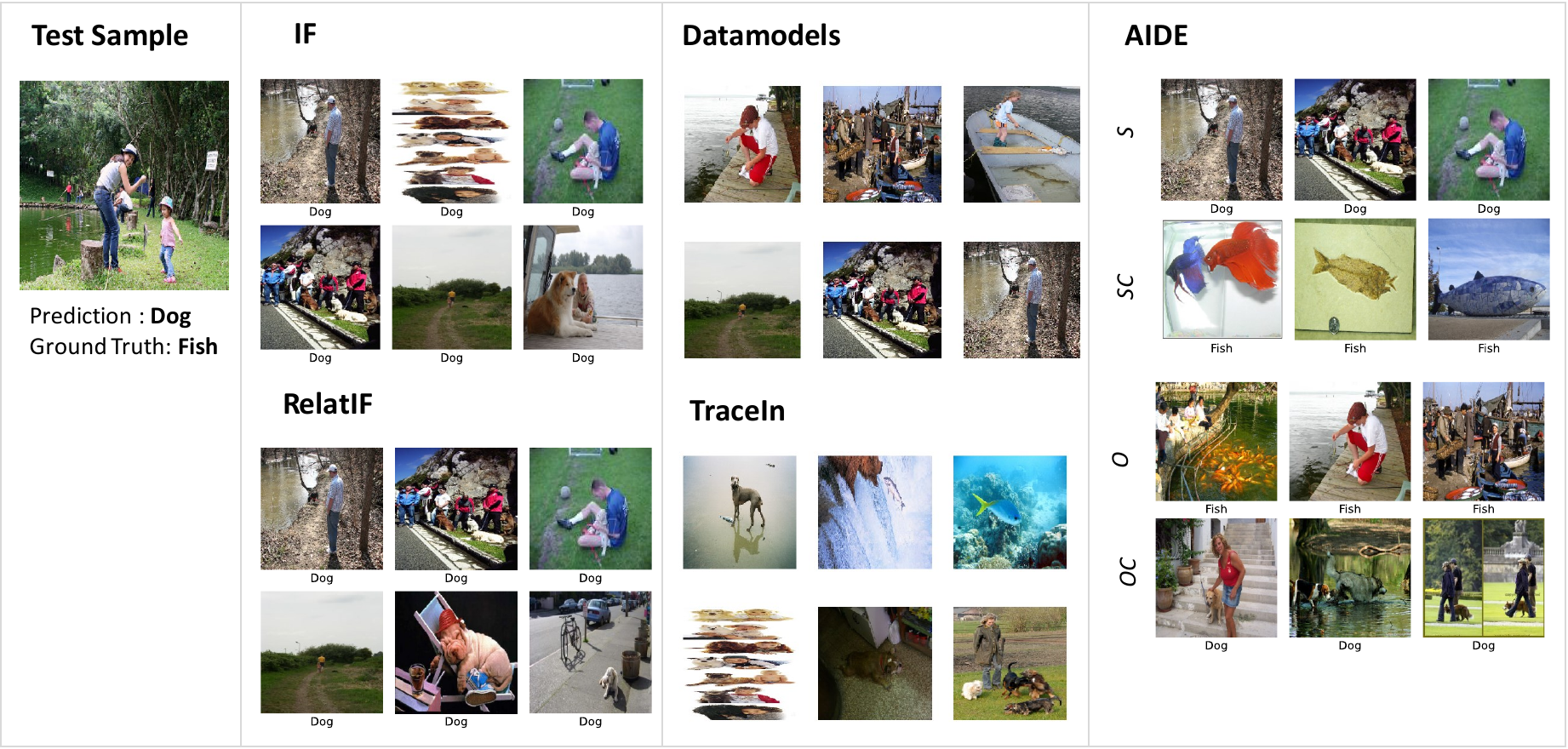}
    \caption{Explanations to investigate a wrong prediction.}
    \label{fig:wrong}
\end{figure}

\begin{figure}[h]
    \centering    \includegraphics[width=0.7\columnwidth]{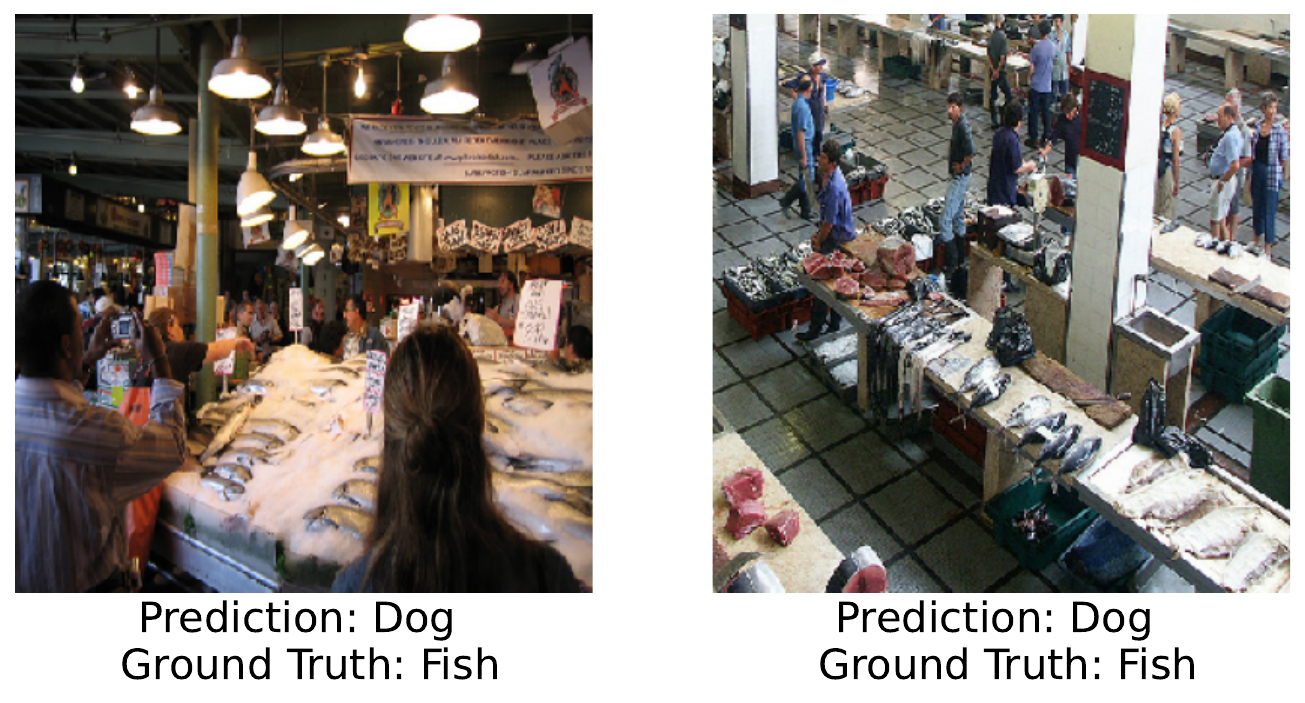}
    \caption{Misclassified test images of fish.}
    \label{fig:miss}
\end{figure}

When examining a wrong prediction for the test sample depicted in Figure \ref{fig:wrong}, where the ground truth label is fish but the model predicted it as a dog, AIDE generates all four sets. After analyzing the S and O, a notable observation is the consistent presence of humans in each example. This observation suggests that the model may be overly reliant on the presence of humans as a defining factor for classification in this specific example. Furthermore, when comparing the S and SC, it becomes apparent that the presence of humans serves as a key distinguishing aspect for the model to classify dogs, which is not the case when classifying fish. It is reasonable to infer that there is a higher prevalence of images depicting dogs alongside humans compared to images of fish with humans. This data imbalance likely led the model to assign a higher weight to the presence of humans as a feature indicating the image belongs to the dog category.
To confirm this, we examined test images where the model's prediction differed from the ground truth and investigated if there was a higher presence of fish images containing humans. As expected, the images in Figure~\ref{fig:miss} were also misclassified due to this factor.
Note that although the explanation of IF can also indicate the importance of humans, it does not comprehensively back up this assumption with contrastivity and by opposers who also contain humans but do not rely on it as much. 

When faced with an ambiguous sample as in Figure \ref{fig:ambig}, where the image contains both a dog and a fish, understanding why the model chose a specific class (in this case, a dog) despite the ground truth being a fish becomes crucial.
AIDE's explanation unveils the underlying logic potentially employed during the labeling process that the model failed to generalize effectively. By examining the S, we observe that the model learns from both dog-related features and water-related features, which aligns with common sense. However, the O suggest the potential existence of a labeling rule that associates images containing both dogs and fish with the ``fish'' label. This rule may not have been strongly represented in the training data, leading to the model's inefficient learning of this specific rule. Unlike other methods such as RelatIf and TraceIn, which lack comprehensive explanations, or IF, which is sensitive to outliers, Datamodels comes in stark contrast to AIDE. 
We observed that when confronted with mislabeled or ambiguous samples, Datamodels may explain the opposite label prediction rather than the model's actual prediction.
This happens due to a discordance between the model being explained and the intermediary models used to compute the importance of individual training examples; in fact, about 20\% of the intermediary models predict a different class that the actual model.

\begin{table}[t]
\scriptsize
\begin{center}
\centering
\caption{AIDE for an ambiguous text message.}
\label{table:ambspam}
\begin{tabular}{p{0.82\linewidth}c}
 \toprule
{\bfseries Test prediction of interest} & {\bfseries Label}\\
\midrule
\textit{`Do you realize that in about 40 years, we'll have thousands of old ladies running around with tattoos?'}
 & \textcolor{red}{Spam} \\
\bottomrule
\toprule
{\bfseries Supporters} &  \\
\midrule
\textit{`Do you ever notice that when you're driving, anyone going slower than you is an idiot and everyone driving faster than you is a maniac?'}
 & \textcolor{red}{Spam} \\
\bottomrule
\textit{`How come it takes so little time for a child who is afraid of the dark to become a teenager who wants to stay out all night? '}
 & \textcolor{red}{Spam}\\
\midrule
\textit{`LIFE has never been this much fun and great until you came in. You made it truly special for me. I won't forget you!'}
  & \textcolor{red}{Spam} \\
\bottomrule
\toprule
{\bfseries Opposers} &  \\
\midrule
\textit{`You are always putting your business out there. You put pictures of your ass on facebook. Why would i think a picture of your room would hurt you, make you feel violated.'}
 & \textcolor{green}{Ham} \\
\midrule
\textit{`Yo you guys ever figure out how much we need for alcohol? Jay and I are trying to figure out how much we can spend on weed'}
 & \textcolor{green}{Ham} \\
\midrule
\textit{`Any chance you might have had with me evaporated as soon as you violated my privacy by stealing my phone number from your employer's paperwork.'}
 & \textcolor{green}{Ham} \\
\bottomrule
\end{tabular}
\end{center}
\end{table}

Table~\ref{table:ambspam} shows another ambiguous test sample for spam classification. Determining whether this message is spam or not is challenging since it does not exhibit the typical form of either a ham message or a common spam message. Instead, it takes the form of an aphorism, which falls into an ambiguous category of messages. AIDE's supporters shed light on the presence of numerous aphorisms in the training set that are labeled as spam, indicating the existence of a labeling logic for categorizing such messages as spam. 
Thus, the model can correctly classify this message despite its ambiguity.
The supporting samples provided by AIDE emphasize a specific logic that was likely injected during the labeling process, indicating that aphorisms were considered spam. These supporting samples contributed to the correct classification decision by reinforcing this logic. 
The opposing examples suggest that classifying the message as non-spam could be a plausible interpretation.
However, the model's ability to correctly classify the message indicates that the rule regarding aphorisms is supported by an adequate number of training samples. This indicates that the model has learned and generalized this rule effectively.

\title{userstudy}
\subsection{User Study}
Following the recommendations of \cite{ustudy}, we invited 33 participants with diverse levels of ML knowledge, including professors, researchers, and PhD students, and non-experts, such as master students; all were non-paid volunteers. Initial inquiries were designed to ascertain participants' knowledge levels, facilitating subsequent segregated assessments of AIDE's effectiveness tailored to different groups. The following questions were asked:
Q: What is your level of familiarity with machine learning concepts?  (Expert, Advanced, Intermediate, Beginner, Unfamiliar)
Q: Are you familiar with the concept of Explainable Artificial Intelligence (XAI)? (Yes, No)

\begin{figure}[h]
    \includegraphics[width=\columnwidth]{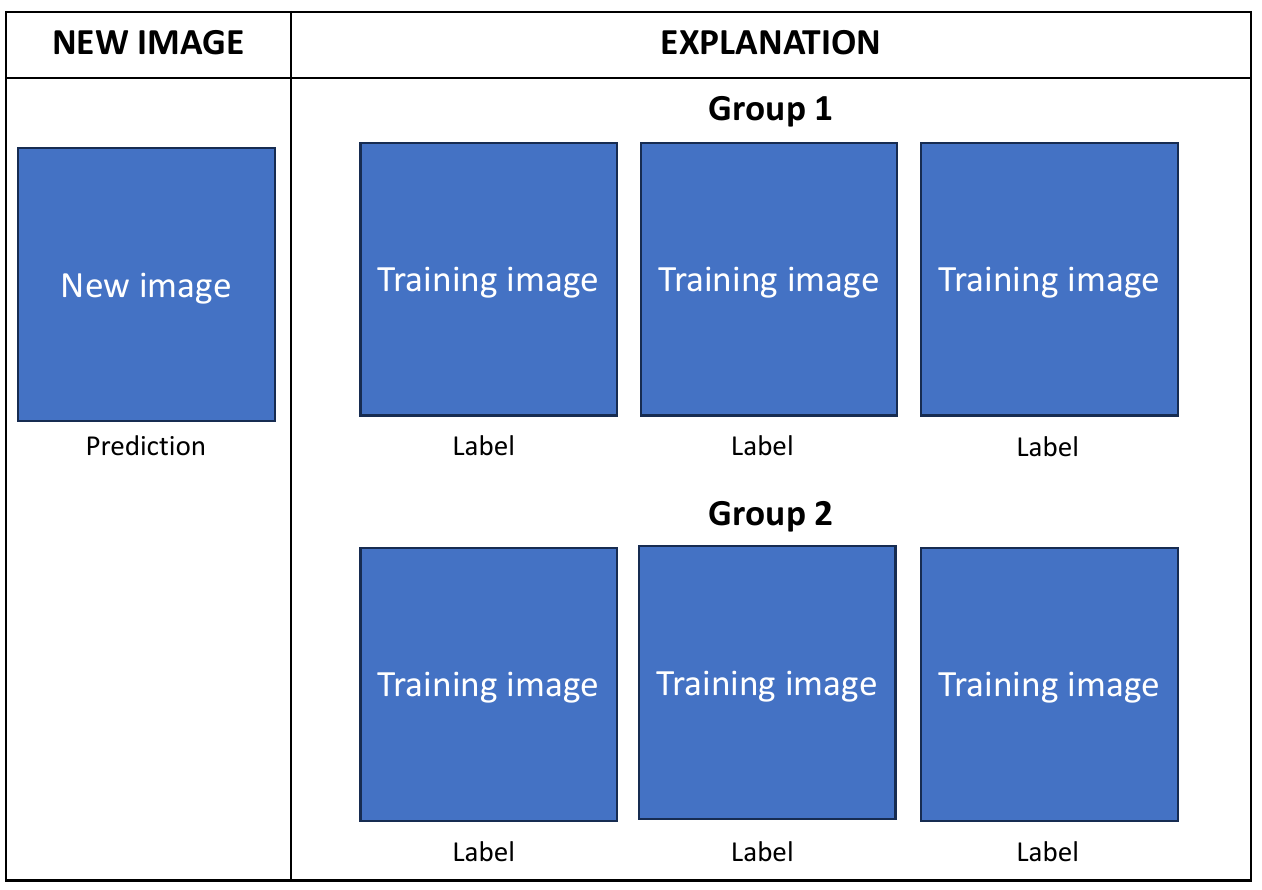}
    \caption{A template of how explanations are presented}
    \label{fig:samplesurvey}
\end{figure}

After knowing their level, we gave a detailed explanation of the dataset, task, and how the explainability technique works. A sample was shown to describe what will be presented, see Figure \ref{fig:samplesurvey}. The user study as it was presented to participants can be found here \footnote{\url{https://forms.gle/AJKe87T9bCrqxsA47}}

The assessment considers:

\textbf{Mental Model}: Q1. The explanation helped to understand the model's prediction. To what extent do you agree?
\textbf{Clarity}: Q2. The explanation is clear and easy to comprehend. To what extent do you agree?
\textbf{Usefulness} of AIDE Quadrants: Q3, Q4, Q5, Q6. The group ``S'', ``SC'', ``O'', ``OC'' enhances understanding the model's prediction. To what extent do you agree?
\textbf{Human-AI Collaboration}: Q7. Did the explanation help understand how the model's performance can be improved?
\textbf{Effectiveness}: Q8. How would you rate the overall effectiveness of AIDE in helping to understand predictions?
\textbf{Helpfulness}: Q9. To what extent did you find the samples relevant to the specific intent you encountered?
\textbf{Contrastivity}: Q10. Do you believe that the use of contrast in the groups of images shown enhanced your understanding of the model predictions?

\begin{figure}[h]
    \centering
    \includegraphics[width=\columnwidth]{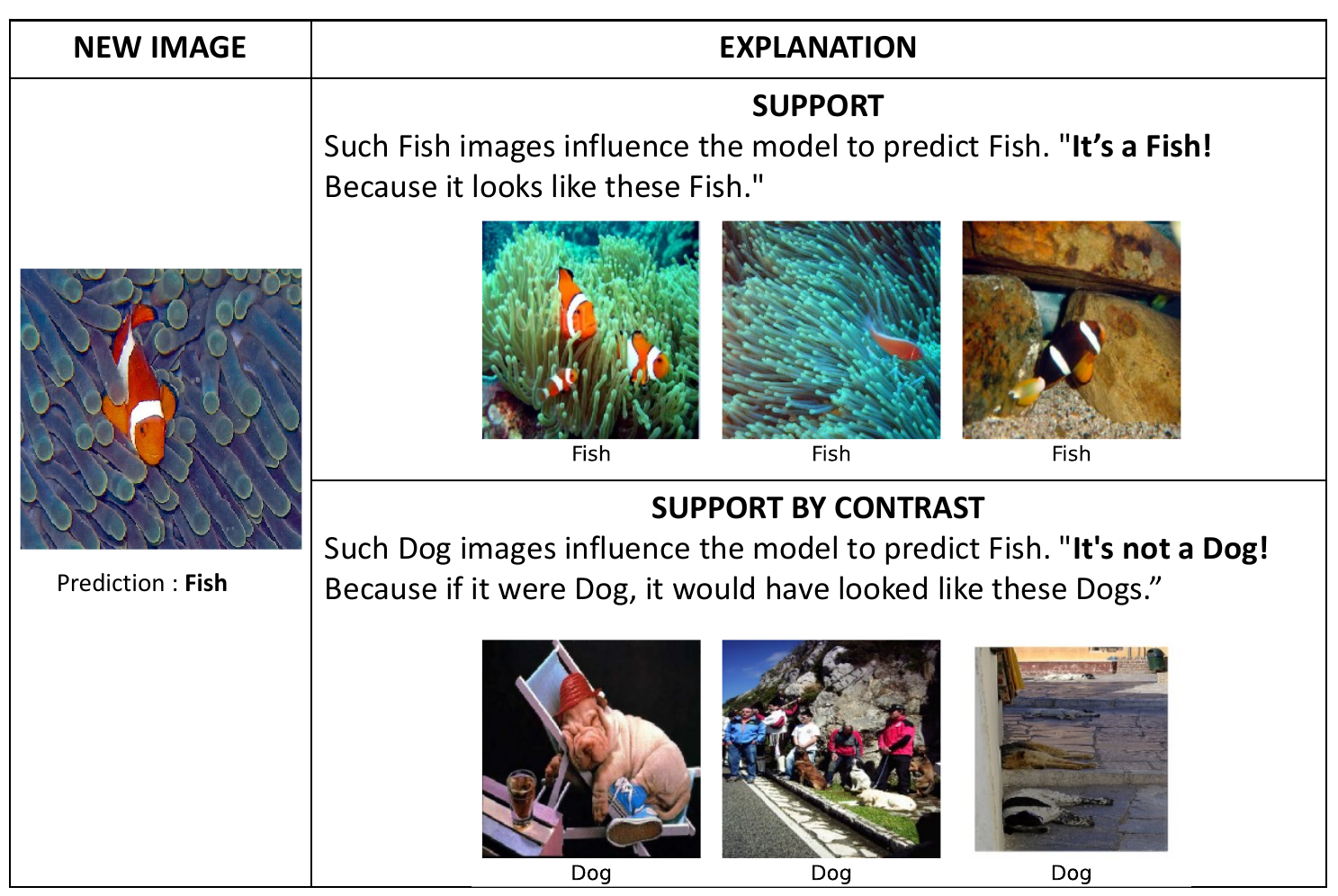}
    \caption{Correct prediction analysis}
    \label{fig:correctsur}
\end{figure}

\begin{figure}
    \centering
    \includegraphics[width=\columnwidth]{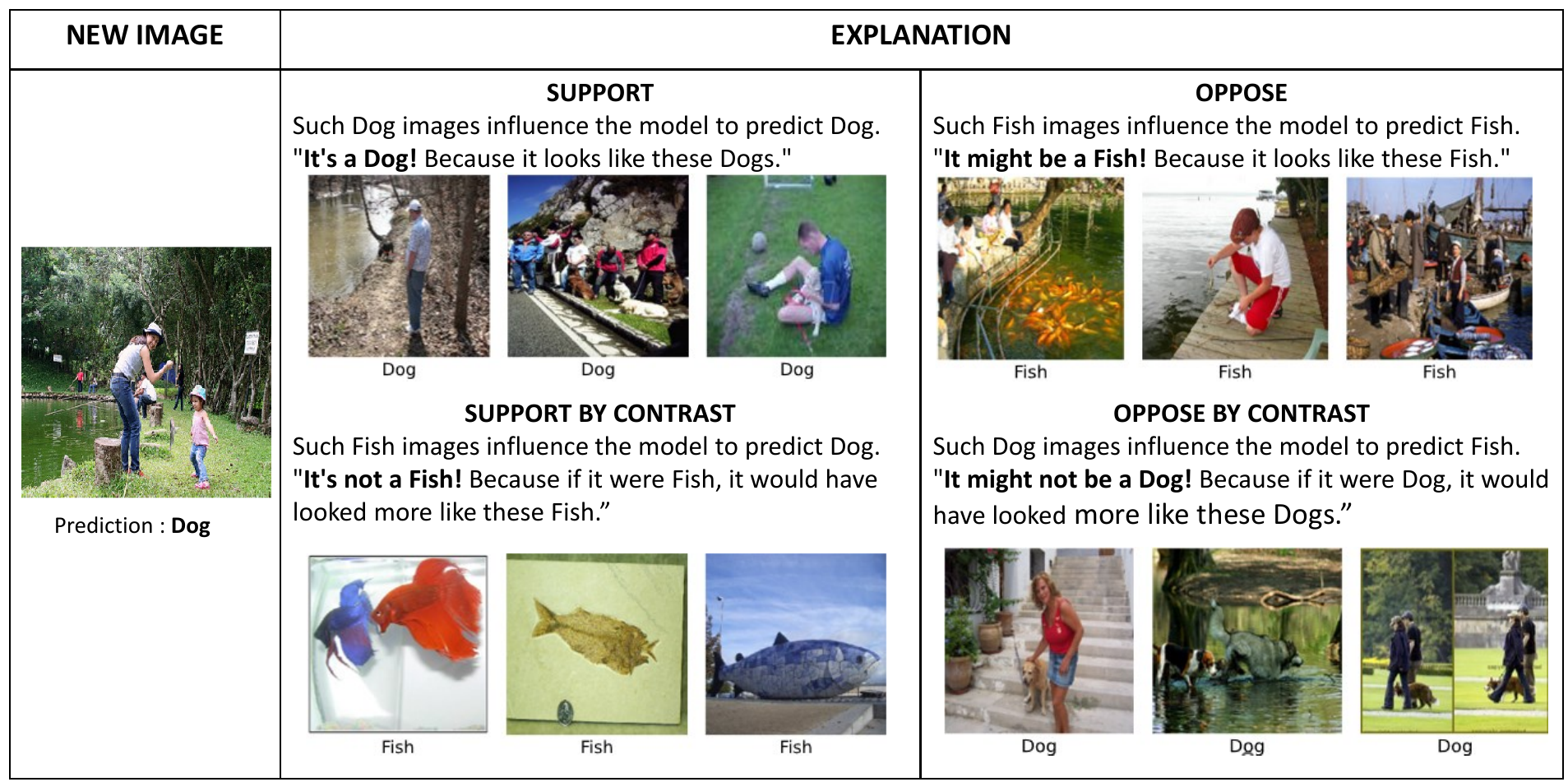}
    \caption{Wrong prediction analysis}
    \label{fig:wrongsur}
\end{figure}

\begin{figure}
    \centering
    \includegraphics[width=\columnwidth]{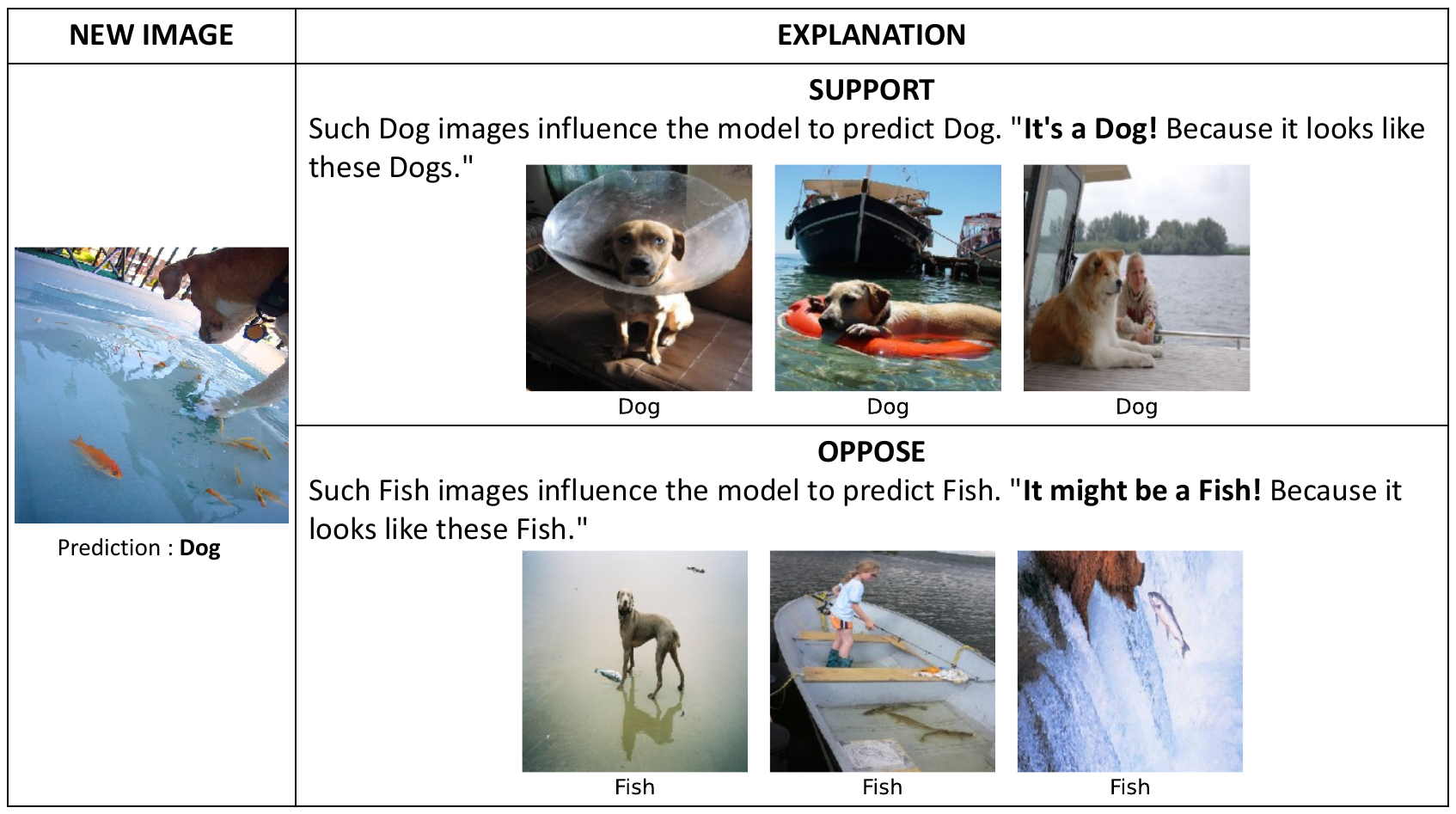}
    \caption{Ambiguous prediction analysis}
    \label{fig:ambsur}
\end{figure}

The evaluation process involved separate analyses of explanations for distinct intents, aiming to gauge the alignment of AIDE with varied user needs. In the forms, we added some extra instructions to better understand what each quadrant wants to say.
The study commenced with the examination of \emph{interpreting a correct prediction}, illustrated in Figure \ref{fig:correctsur}. Subsequently, for the investigation of a wrong prediction and clarification of an ambiguous one, Figures \ref{fig:wrongsur} and \ref{fig:ambsur} were presented, respectively. Sequentially, the questions outlined in the main paper were posed for each intent. The study concludes with general inquiries to elicit an overarching evaluation of the approach.

In Table \ref{table:us_int} and Table \ref{table:us_ev}, the results are presented through two distinct aggregation approaches. Firstly, responses from participants are combined based on their expertise level, forming two groups: ``advanced'', which adds up responses from experts and advanced participants, and ``intermediate'', which combines responses from beginner and intermediate participants. The second aggregation method involves the transformation of a 5-point Likert scale into a simplified two-point scale. For instance, responses such as ``Agree'' and ``Somewhat agree'' are merged into the category of ``Agree'', while responses including ``Disagree'', ``Somewhat disagree'', and ``Neutral'' are consolidated into the category of ``Disagree''. The subsequent discussion focuses on the results without these aggregation techniques and question by question.

Out of 33 participants, the distribution across expertise levels was 3 experts, 13 advanced, 12 intermediate, and 5 beginners. The possible answers were given in a 5-point Likert scale, denoted as follows:
\textcolor[HTML]{32CB00}{\textbf{OO}} -- Strongly agree;
\textcolor[HTML]{32CB00}{\textbf{O}} -- Somewhat agree;
N -- Neither agree nor disagree;
\textcolor[HTML]{FE0000}{\textbf{X}} -- Somewhat disagree;
\textcolor[HTML]{FE0000}{\textbf{XX}} -- Disagree.

In Table \ref{table:us_int}, the percentage of participants who agreed on the high quality of specific aspects of AIDEs' explanation for particular intents is presented. Whereas, in Table \ref{table:us_ev}, the percentages of users who overall highly assessed AIDE's effectiveness, the utility of contrast in explanation, and AIDE's capability to tailor explanations according to user intent are depicted. Notably, participants with more advanced expertise tend to rate highly more frequently across various aspects of AIDE's explanation. We also note that a positive response to the intent-based nature of explanations (Q7) can facilitate improved human-XAI collaboration, which is currently suboptimal \cite{aies3}.
\begin{table}[]
\caption{Percentage (\%) of people who agree with Q1--7.}
\label{table:us_int}
\resizebox{\columnwidth}{!}{%
\begin{tabular}{|c|c|c|c|c|}
\hline
\textbf{}            & \textbf{ML knowl.} & \textbf{Int. correct} & \textbf{Inv. wrong} & \textbf{Cl. ambiguous} \\ \hline
\multirow{2}{*}{\textbf{Q1}} & Advanced              & 88                    & 88                  & 69                     \\ \cline{2-5} 
                             & Intermediate          & 87                    & 67                  & 73                     \\ \hline
\multirow{2}{*}{\textbf{Q2}} & Advanced              & 94                    & 81                  & 81                     \\ \cline{2-5} 
                             & Intermediate          & 87                    & 66                  & 73                     \\ \hline
\multirow{2}{*}{\textbf{Q3}} & Advanced              & 100                   & 88                  & 88                     \\ \cline{2-5} 
                             & Intermediate          & 93                    & 80                  & 80                     \\ \hline
\multirow{2}{*}{\textbf{Q4}} & Advanced              & 75                    & 81                  & -                      \\ \cline{2-5} 
                             & Intermediate          & 67                    & 66                  & -                      \\ \hline
\multirow{2}{*}{\textbf{Q5}} & Advanced              & -                     & 63                  & 60                     \\ \cline{2-5} 
                             & Intermediate          & -                     & 80                  & 73                     \\ \hline
\multirow{2}{*}{\textbf{Q6}} & Advanced              & -                     & 63                  & -                      \\ \cline{2-5} 
                             & Intermediate          & -                     & 66                  & -                      \\ \hline
\multirow{2}{*}{\textbf{Q7}} & Advanced              & -                     & 88                  & 69                     \\ \cline{2-5} 
                             & Intermediate          & -                     & 87                  & 67                     \\ \hline
\end{tabular}
}
\end{table}

\begin{table}[]
\caption{Percentage (\%) of people who agree with Q8--10.}
\label{table:us_ev}
\centering
\footnotesize
\begin{tabular}{|c|l|c|c|}
\hline
\multicolumn{1}{|l|}{\textbf{ ML knowledge}}& \textbf{Q8} & \multicolumn{1}{l|}{\textbf{Q9 }} & \multicolumn{1}{l|}{\textbf{Q10}} \\ \hline
Advanced     &88  &100  & 100 \\ \hline
Intermediate &80  & 73 & 100 \\ \hline
\end{tabular}%
\end{table}

Note that we do not compare AIDE with other methods to avoid \emph{participant bias}, where the participant's behavior is affected once they deduce what the preferred answers of the researcher are. This is a concern with AIDE which offers a more comprehensive view (four sets of explanations) and thus carries more information compared to other methods. Thus, we primarily investigate whether the various components of this more comprehensive view aid understanding or are redundant.
Specifically, we implicitly draw conclusions on the added value of AIDE, by assessing: (1) the \emph{significance of the other three quadrants} (Q4, Q5, Q6), where 63\%--81\% of participants agree; (2) \emph{intent nature of explanations}, where 87\% of participants liked; and (3) the \emph{importance of contrastivity} (Q10), where 100\%  agree.

Other findings from the user study can be found in the following tables.

\begin{table}[h]
\centering
\begin{tabular}{|l|l|l|l|l|l|l|}
\hline
\textbf{Intent} &
  \textbf{Level} &
  {\color[HTML]{32CB00} \textbf{OO}} &
  {\color[HTML]{32CB00} \textbf{O}} &
  \textbf{N} &
  {\color[HTML]{FE0000} \textbf{X}} &
  {\color[HTML]{FE0000} \textbf{XX}} \\ \hline
                                & Expert       & 1          & 2 &   &   &   \\ \cline{2-7} 
                                & Advanced     & 8          & 5          &   &   &   \\ \cline{2-7} 
                                & Intermediate & 8          & 2          & 1 & 1 &   \\ \cline{2-7} 
\multirow{-4}{*}{Int. correct}  & Beginner     & 2          & 3          &   &   &   \\ \hline
                                & Expert       & 1          & 2          &   &   &   \\ \cline{2-7} 
                                & Advanced     & 4          & 7          & 1 & 1 &   \\ \cline{2-7} 
                                & Intermediate & 4          & 3          & 2 & 3 &   \\ \cline{2-7} 
\multirow{-4}{*}{Inv. wrong}    & Beginner     & 1           & 4          &   &   &   \\ \hline
                                & Expert       & 1          & 1          &   &   & 1 \\ \cline{2-7} 
                                & Advanced     & 7          & 2          & 3 & 1 &   \\ \cline{2-7} 
                                & Intermediate & 4          & 5          & 1 & 2 &   \\ \cline{2-7} 
\multirow{-4}{*}{Cl. ambiguous} & Beginner     & 2          & 2          &   &   & 1 \\ \hline
\end{tabular}%
\caption{\textbf{Mental Model}: Q1 -- The explanation provided helped to understand the model's prediction. To what extent do you agree?}
\end{table}

\begin{table}[h]
\centering
\begin{tabular}{|l|l|l|l|l|l|l|}
\hline
\textbf{Intent} &
  \textbf{Level} &
  {\color[HTML]{32CB00} \textbf{OO}} &
  {\color[HTML]{32CB00} \textbf{O}} &
  \textbf{N} &
  {\color[HTML]{FE0000} \textbf{X}} &
  {\color[HTML]{FE0000} \textbf{XX}} \\ \hline
                                & Expert       & 2          & 1 &   &   &   \\ \cline{2-7} 
                                & Advanced     & 9          & 3          &   &   &1   \\ \cline{2-7} 
                                & Intermediate & 7          & 3          &  & 2 &   \\ \cline{2-7} 
\multirow{-4}{*}{Int. correct}  & Beginner     & 2          & 3          &   &   &   \\ \hline
                                & Expert       &           & 2          &   &1   &   \\ \cline{2-7} 
                                & Advanced     & 4          & 7          & 1 & 1 &   \\ \cline{2-7} 
                                & Intermediate & 4          & 3          &2  & 3 &   \\ \cline{2-7} 
\multirow{-4}{*}{Inv. wrong}    & Beginner     & 2           & 3          &   &   &   \\ \hline
                                & Expert       & 2          &           &   &   & 1 \\ \cline{2-7} 
                                & Advanced     & 7          & 4          &2 &  &   \\ \cline{2-7} 
                                & Intermediate & 4          & 5          & 1 & 2 &   \\ \cline{2-7} 
\multirow{-4}{*}{Cl. ambiguous} & Beginner     & 2          & 2          &   &1   &  \\ \hline
\end{tabular}%
\caption{\textbf{Clarity}: Q2 -- The explanation is clear and easy to comprehend. To what extent do you agree? }
\end{table}

\begin{table}[h]
\centering
\begin{tabular}{|l|l|l|l|l|l|l|}
\hline
\textbf{Question} &
  \textbf{Level} &
  {\color[HTML]{32CB00} \textbf{OO}} &
  {\color[HTML]{32CB00} \textbf{O}} &
  \textbf{N} &
  {\color[HTML]{FE0000} \textbf{X}} &
  {\color[HTML]{FE0000} \textbf{XX}} \\ \hline
                                & Expert       &  4          & 3 &1   &1   &   \\ \cline{2-7} 
                                & Advanced     & 23          & 14          &1   &   &1   \\ \cline{2-7} 
                                & Intermediate & 22          & 9          &2  & 3 &   \\ \cline{2-7} 
\multirow{-4}{*}{Q3}  & Beginner     & 8          & 6          &   &1   &1   \\ \hline
                                & Expert       &2           & 2          &   &2   &   \\ \cline{2-7} 
                                & Advanced     & 13          & 8          & 3 & 1 &1   \\ \cline{2-7} 
                                & Intermediate & 12          & 4          &2  & 5 &1   \\ \cline{2-7} 
\multirow{-4}{*}{Q4}    & Beginner     & 5           & 3          &1   &1   &   \\ \hline
                                & Expert       & 2          &1           &1   &2   &  \\ \cline{2-7} 
                                & Advanced     & 7          & 8          &7 &4  &   \\ \cline{2-7} 
                                & Intermediate & 10          & 8          & 2 & 4 &   \\ \cline{2-7} 
\multirow{-4}{*}{Q5} & Beginner     & 3          & 6          &   &1   &  \\ \hline
                                & Expert       &           & 2          &   &1   &  \\ \cline{2-7} 
                                & Advanced     & 3          &5          &4 &  &1   \\ \cline{2-7} 
                                & Intermediate & 4          & 5          & 1 & 2 &   \\ \cline{2-7} 
\multirow{-4}{*}{Q6} & Beginner     & 2          & 6          &1   &3   &  \\ \hline
\end{tabular}%
\caption{\textbf{Usefulness} of AIDE Quadrants: Q3/Q4/Q5/Q6 -- The group ``Support''/``Support by Contrast''/``Oppose''/``Oppose by Contrast'' enhances the understanding of the model's prediction. To what extent do you agree? }
\end{table}

\begin{table}[h]
\centering
\begin{tabular}{|l|l|l|l|}
\hline
\textbf{Intent} &
  \textbf{Level} &
  \textbf{Yes} &
  \textbf{No} \\ \hline
                                & Expert       & 1          & 1  \\ \cline{2-4} 
                                & Advanced     & 12          & 1 \\ \cline{2-4} 
                                & Intermediate & 11          & 1 \\ \cline{2-4} 
\multirow{-4}{*}{Inv. wrong}    & Beginner     & 4           & 1 \\ \hline
                                & Expert       & 2          & 1    \\ \cline{2-4} 
                                & Advanced     & 9          & 4     \\ \cline{2-4} 
                                & Intermediate & 9          & 3     \\ \cline{2-4} 
\multirow{-4}{*}{Cl. ambiguous} & Beginner     & 3          & 2  \\ \hline
\end{tabular}%
\caption{\textbf{Human-AI Collaboration}: Q7 -- Did the explanation help understand how the model's performance can be improved? }
\end{table}

\begin{table}[h]
\centering
\begin{tabular}{|l|l|l|l|l|l|l|}
\hline
\textbf{Question} &
  \textbf{Level} &
  {\color[HTML]{32CB00} \textbf{OO}} &
  {\color[HTML]{32CB00} \textbf{O}} &
  \textbf{N} &
  {\color[HTML]{FE0000} \textbf{X}} &
  {\color[HTML]{FE0000} \textbf{XX}} \\ \hline
                                & Expert       & 1          & 2 &   &   &   \\ \cline{2-7} 
                                & Advanced     &4          & 7          &2   &   &   \\ \cline{2-7} 
                                & Intermediate & 2          & 8          &1  & 1 &   \\ \cline{2-7} 
\multirow{-4}{*}{Q8}  & Beginner     &2          & 3          &   &   &   \\ \hline
                                & Expert       &           & 3          &   &   &   \\ \cline{2-7} 
                                & Advanced     & 6          & 7          &  &  &   \\ \cline{2-7} 
                                & Intermediate & 4          & 6          &2  &  &   \\ \cline{2-7} 
\multirow{-4}{*}{Q9}    & Beginner     & 1          & 2          &2   &   &   \\ \hline
                                & Expert       & 1          &2           &   &   &  \\ \cline{2-7} 
                                & Advanced     & 9          & 4          & &  &   \\ \cline{2-7} 
                                & Intermediate & 3          & 8          & 1 &  &   \\ \cline{2-7} 
\multirow{-4}{*}{Q10} & Beginner     & 4          & 1          &   &   &  \\ \hline
\end{tabular}%
\caption{\textbf{Effectiveness}: Q8 -- How would you rate the overall effectiveness of AIDE in helping to understand model predictions?; 
\textbf{Helpfulness}: Q9 -- To what extent did you find the provided samples relevant to the specific intent you encountered?; 
\textbf{Contrastivity}: Q10 -- Do you believe that the use of contrast in the groups of images shown enhanced your understanding of the model predictions?}
\end{table}

\section{Conclusion}
\label{sec:conclusion}
In this paper, we introduce AIDE, a novel example-based explainability method that generates diverse and contrastive explanations tailored to user's needs and intentions. Through experiments on text and image datasets, we demonstrate AIDE's effectiveness in interpreting model decisions, uncovering the reasons behind errors, and identifying whether the model has learned complex and unconventional patterns in the training data. Quantitative and qualitative analysis affirms that AIDE outperforms existing approaches.
\section{Limitations}
One limitation is that we exclusively tested the approach on binary classification. Extending our work beyond binary classification mostly concerns the "opposite label-same label" dimension. There are two options: (a) one-vs-rest, and group all opposite labels together similar to the "by contrast" quadrants, or (b) one-vs-one, and break down opposite labels. This impacts presentation where the latter approach requires users to build a bit more complex mental model.

Another limitation is our user study's relatively small sample size, comprising only 33 participants with diverse backgrounds. This limited sample size precluded the possibility of conducting a large-scale and extensive survey, where we could include examples from existing methods for a control group not exposed to the AIDE explanations, thus mitigating the bias discussed above. By integrating results from both groups, we could more robustly demonstrate the superiority of our methods compared to the baselines.
\section*{Acknowledgments}
This work was supported by the European Union under Horizon 2020 grant agreement No. 955895.

\newpage
\bibliography{aaai24}

\end{document}